\documentclass[lettersize,journal]{IEEEtran}
\usepackage{amsmath,amsfonts,amssymb}
\usepackage{algorithm}
\usepackage{algorithmic}
\usepackage{array}
\usepackage[caption=false,font=normalsize,labelfont=sf,textfont=sf]{subfig}
\usepackage{textcomp}
\usepackage{stfloats}
\usepackage{url}
\usepackage{verbatim}
\usepackage{graphicx}
\usepackage{cite}
\usepackage{xcolor}
\usepackage{siunitx}
\usepackage[normalem]{ulem}
\usepackage{gensymb}
\usepackage{xcolor}
\usepackage{soul}
\usepackage[colorlinks=true,linkcolor=black,citecolor=blue,urlcolor=blue]{hyperref}
\usepackage{bookmark}

\begin{document}

\title{Task-space model-based control of pneumatic soft actuators}

\author{Nithin S. Kumar$^{1}$, Joshua Gaston$^{2}$, D. Caleb Rucker$^{2}$, Eric J. Barth$^{1}$

\thanks{$^{1}$ Department of Mechanical Engineering,
        Vanderbilt University, Nashville, Tennessee 37235. Email: nithin.s.kumar@vanderbilt.edu 
        }%
\thanks{$^{2}$ Department of Mechanical, Aerospace, and Biomedical Engineering, The University of Tennessee, Knoxville, TN 37996}        
        \thanks{This work has been submitted to the IEEE for possible publication. Copyright may be transferred without notice, after which this version may no longer be accessible.}
        }

\maketitle
\begin{abstract}
Soft actuators enable dexterous and compliant interaction, but closed-loop task-space control remains challenging due to strong nonlinearities, distributed deformation, and uncertainty in their dynamics. This paper presents a real-time dynamic-model-based task-space feedback and estimation framework based on a non-minimal coordinate discrete elastic rod model formulated in absolute coordinates with holonomic constraints. The resulting structure preserves distributed mechanics while maintaining computational efficiency through sparse system matrices, enabling real-time control with up to 10 discretized rods. A quasi-static feedforward inverse model is combined with a task-space PI controller and a dynamic observer that fuses measurement residuals as virtual forces, enabling full-state estimation from sparse sensing. The approach is experimentally validated on three planar pneumatic soft actuators with varying geometries. Across five tasks—including drawing the digits 0-9 across the workspace (3–18 mm/s tip speed), tracking periodic motion (up to 37 cm/s), cross-platform generalization, reduced sensing conditions, and real-time user-defined references—our method achieves 1.5–2.3 mm root mean square error (RMSE) for precision motions and 5.5–12.4 mm RMSE at 1–2 Hz. Results demonstrate that structured, non-minimal dynamic models can enable real-time, high-precision, moderate-bandwidth task-space control of planar soft pneumatic actuators in free space.
\end{abstract}


\section{Introduction}
Soft continuum actuators offer advantages such as dexterity, passive compliance, and safe interaction for manipulation, medical procedures, and operation in confined environments \cite{Burgner-Kahrs2015,DupontSurvey2022,Laschi2016}. Soft pneumatic actuators (e.g., Pneunets \cite{ilievski2011soft}) are particularly attractive because of their low cost, simple fabrication, and ability to conform to complex surroundings \cite{El-Atab2020,Su2022,Xavier2022b}. Achieving dynamic task-space performance with these systems, however, remains challenging: they are underactuated, highly deformable, and effectively infinite-dimensional. Therefore, controllers must compensate for model uncertainty, nonlinear distributed elasticity, and pneumatic bandwidth limitations while remaining computationally efficient for real-time implementation.

Early continuum-robot controllers often used kinematic end-effector regulation with Jacobian-based inverse kinematics and classical P/PD/PID feedback~\cite{Bailly2005,Camarillo2009P,Mahvash2011,DupontDesignTRO10}. While effective for quasi-static motion, these approaches generally require conservative gains and do not explicitly incorporate full manipulator inertia, limiting achievable bandwidth.

Dynamic controllers typically use a model-based reduced-order approach \cite{DellaSantina2018,Xavier2022,Azizkhani2025}, often augmented with learning components \cite{Braganza2007,Gillespie2018,Haggerty2023}. These reduced-order models are often minimal-coordinate, relative-geometry models and assume piecewise-constant curvature (PCC) \cite{Webster2010a,Godage2016,Katzschmann2019}, which discretizes the actuator into segments with finite constant curves. While PCC models are effective for quasi-static conditions and negligible external loading, they are unable to capture higher-order vibrations, curvature variations, and distributed dynamic effects unless discretized finely. However, their dense inertia matrices and cubic scaling in computation time \cite{Till2019,Boyer2021} make this intractable for real-time operation beyond a few segments (e.g., three to five segments in \cite{Kapadia2014,Azizkhani2025,Falkenhahn2017,DellaSantina2018}). Learning-based controllers can improve performance but introduce training overhead, lack generalizability, and offer limited stability guarantees \cite{Thuruthel2019,Haggerty2023}.

\begin{figure}
  \centering
  \includegraphics[width=0.75\linewidth]{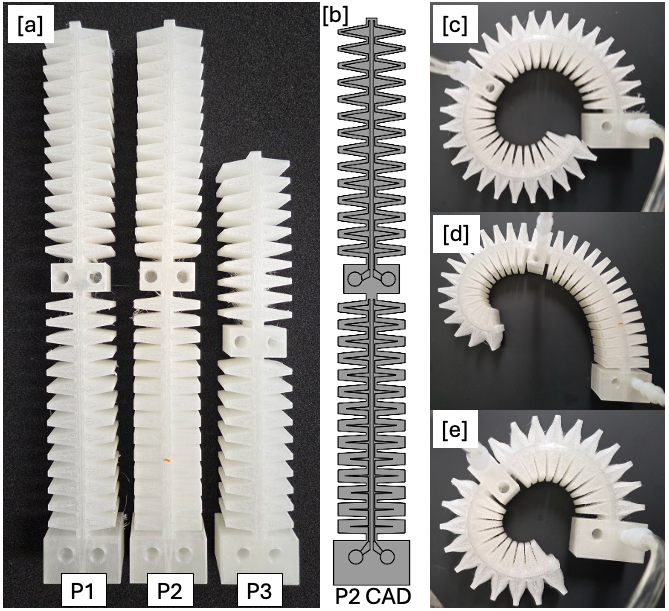}
  \caption{(a) The three planar soft pneumatic actuators used in this study consist of four chambers each. Tasks 1, 2, 4, and 5 are conducted with P1; Task 3 evaluates generalization using P2 and P3. P1 and P3 have the same constant cross-section and differ only in length, while the wall thickness of P2 decreases from base to tip as shown in a CAD cross-sectional view (b). Two chambers of the same side of each actuator are pressurized to 35 psi (241 kPa) resulting in bending for (c) P1, (d) P2, and (e) P3.}
  \label{fig:pneunet}
\end{figure}

An alternative modeling direction is to adopt discrete elastic rod formulations that preserve distributed mechanics without imposing curvature shape assumptions. However, such models are typically expressed in minimal or reduced coordinates and are often viewed as impractical for real-time control due to the presence of algebraic constraints and increased state dimension (e.g., \cite{Renda2020}). In contrast, we demonstrate that a non-minimal, constraint-based discrete rod formulation can be implemented efficiently and remains computationally tractable for real-time task-space control at higher spatial resolutions.

In particular, we apply this modeling framework to control the tip position of a planar pneumatic soft actuator (denoted P1 in Fig. \ref{fig:pneunet}a). The model builds on prior discrete rod and constrained Lagrangian formulations \cite{Rucker2022,Kim2026} and provides several advantages:

\begin{itemize}
\item \textit{Computational efficiency with high spatial resolution:} The resulting equations exhibit a sparse, block tri-diagonal structure \cite{Till2019,Gaston2026}, enabling real-time control with up to 10 discretized rod elements without forming dense inertia matrices.
\item \textit{Direct incorporation of constraints and interactions:} Holonomic and non-holonomic constraints, as well as task-space interaction forces, are embedded directly within the constrained dynamics framework \cite{Kumar2026}.
\item \textit{Observer-compatible structure:} The formulation supports a force-based dynamic observer \cite{Gaston2026} that reconstructs full-state estimates from sparse sensing (e.g., tip-only measurements).
\end{itemize}

The proposed control architecture combines a quasi-static model-based feedforward term with a task-space PI controller and an online dynamic observer. To evaluate performance and versatility, we validate the approach across five representative tasks: 
\begin{enumerate}
    \item Precision trajectory tracking (3-18 mm/s), 
    \item High-speed periodic motion (37 cm/s, 4.7 m/s$^2$),
    \item Generalization to actuators with differing geometries (P2 and P3 in Fig. \ref{fig:pneunet}a),
    \item Trajectory tracking with intermediate-point sensing only,
    \item Real-time user-defined reference tracking.
\end{enumerate}
These tasks were selected to evaluate complementary capabilities needed for practical free-space soft-actuator deployment: precision, bandwidth, cross-platform model generalization, reduced-sensing operation, and online adaptability. To our knowledge, this is the first experimental demonstration of real-time task-space control using a dynamic constrained-rod observer and model-based feedback architecture for planar soft pneumatic continuum actuators across this range of tasks within a single model-based control and estimation framework implemented on physical hardware. Furthermore, this work demonstrates that structured, non-minimal coordinate dynamic models can support real-time planar free-space task-space control at resolutions previously considered computationally impractical.

\section{Methods}

\subsection{Constrained Dynamic Model}
\begin{figure}
  \centering
  \includegraphics[width=0.75\linewidth]{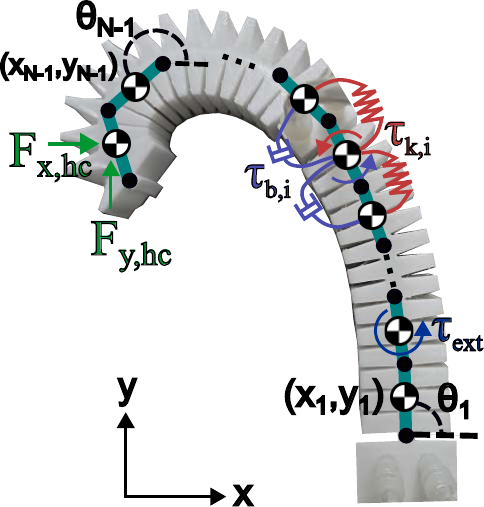}
  \caption{The planar discretized soft actuator model is overlaid with the physical actuator (P2). The actuator backbone is represented by $N$ rigid rods with rotational spring–damper elements generating elastic ($\tau_k$) and damping ($\tau_b$) torques. Adjacent rods are connected through holonomic constraint forces ($F_{x,hc},F_{y,hc}$). Pneumatic actuation is modeled as generalized torques ($\tau_{ext}$) applied at selected rod indices.}
  \label{fig:model}
\end{figure}

The backbone of the soft continuum actuator is discretized as a chain of $N$ rigid rods with rotational elasticity and damping as shown in Fig. \ref{fig:model}. A non-minimal (maximal) set of generalized coordinates is used to describe the configuration:
\begin{equation}
    q = [x_1, y_1, \theta_1, ..., x_N, y_N, \theta_N]^T,
\end{equation}
where ($x_i,y_i$) and $\theta_i$ denote the absolute center-of-mass position and orientation of rod $i$, respectively. Adjacent rods are connected through holonomic constraints enforcing geometric continuity, and the base of rod 1 is pinned to the origin. These constraints are written as $\phi_H(q)=0$, with Jacobian $A = \frac{\partial \phi_H}{\partial q}$. Using a constrained Lagrangian formulation, the equations of motion are:
\begin{gather}
    \frac{d}{dt}\frac{\partial L}{\partial \dot{q}} - \frac{\partial L}{\partial q} = Q_{NC} + A^T(q)\lambda_{H}, \\
    \phi_H(q) = 0,
\end{gather}
where $\lambda_H$ are Lagrange multipliers associated with the holonomic constraints. To mitigate numerical drift during integration, Baumgarte stabilization \cite{baumgarte} is applied by differentiating the constraints to the acceleration level and imposing second-order constraint dynamics. The resulting index-1 differential-algebraic system is:

\begin{align}
\label{eq:fullsystem}
\begin{bmatrix}
M & A^T \\
A & 0 
\end{bmatrix}
\begin{bmatrix}
\Ddot{q} \\
-\lambda_H 
\end{bmatrix}
=
\begin{bmatrix}
Q \\
-\dot{A}\dot{q} - 2\zeta_B \omega_B A \dot{q} - \omega_B^2 \phi_H 
\end{bmatrix},
\end{align}
where $M$ is the diagonal inertia matrix, $\zeta_B$ and $\omega_B$ regulate constraint stabilization dynamics, and $Q$ collects all internal and external generalized forces. 

Internal viscoelastic torques ($\tau_k,\tau_b$ in Fig. \ref{fig:model}) arise from relative angular displacements and velocities between adjacent rods and are modeled as linear rotational springs and dampers, yielding block-tridiagonal stiffness and damping contributions in $Q$. Pneumatic actuation is modeled as equivalent external generalized torques ($\tau_{ext}$ in Fig. \ref{fig:model}) applied at discrete rod indices. The resulting constrained system preserves a sparse, block-structured form that enables efficient linear solves at each integration step. A complete derivation of the constrained Lagrangian formulation, including non-holonomic extensions, is provided in \cite{Kumar2026}.

\subsection{Parameter Estimation}

The discrete rod model is parameterized by rod mass ($m_i$), length ($L_i$), and the rotational stiffness ($k_i$) and damping ($b_i$) constants. Geometric and mass properties ($m_i, L_i$) were obtained by direct measurement. The rotational stiffness parameter ($k_i$) was identified by pressurizing individual chambers and performing a least-squares fit to the measured static deflections, following a procedure similar to \cite{Katzschmann2019}. By construction, the inertia matrix is diagonal, $M=\text{diag}(m_i,m_i,\frac{m_i L_i^2}{12})\in \mathbb{R}^{3N \times 3N}$, where each rod is modeled as a uniform slender body. The stiffness matrix $K$ is block tridiagonal, where each internal rotational DOF contributes a term of the form $-k_i(2\theta_i - \theta_{i-1} - \theta_{i+1})$.

The damping constant ($b_i$) was estimated from the measured free oscillation of the actuator tip using logarithmic decrement to obtain the dominant modal damping ratio $\zeta$. A Rayleigh damping model was assumed, $B=\alpha M + \beta K$, with $(\alpha,\beta)$ selected to bound the damping ratios within a narrow range around the measured $\zeta$, similar to \cite{Katzschmann2019v2,Huang2020}. This prevents excessive damping of higher-frequency modes, which would otherwise introduce numerical stiffness and prohibit fixed-step integration at practical time steps (e.g., $\Delta t = 10^{-3}$ s). The resulting damping model preserves the observed decay characteristics while maintaining computational efficiency. 

Model parameters for the three soft actuators are given in Table \ref{tab:parameters}. Pressurization of a chamber induces opposing bending moments at its boundaries, which are reduced to net torques applied at the proximal, distal, and mid-chamber rod indices. These torques enter the model through the generalized force vector $Q$ as in \cite{Kumar2026}.

\subsection{Controller Architecture}

The overall control structure is shown in Fig.~\ref{fig:control}. A quasi-static model-based feedforward term is combined with a task-space PI controller and a dynamic observer to generate the input vector $u=[u_1,u_2]^T \in \mathbb{R}^{2\times1}$. Here, $u$ represents differential actuation commands applied to opposing chamber pairs of the Pneunet. In particular, $u_1$ actuates the rear pair of chambers: a positive value corresponds to pressurizing one chamber while the opposing chamber remains unpressurized, whereas a negative value reverses this assignment. The second input $u_2$ acts similarly on the top chamber pair. 

\begin{table}
\begin{center}
\caption{Model Parameters}
\label{tab:parameters}
\begin{tabular}{| c | c | c | c |}
\hline
Parameter & P1 & P2 & P3 \\
\hline
Mass [g] & 125.0 & 144.5 & 92.2 \\
\hline 
Length [mm] & 253 & 253 & 193 \\
\hline 
$k_i$ [mNm/rad] & 75.1 & 24.8-66.4 & 99.8 \\
\hline 
$b_i$ [mNm/rad/s] & 0.167 & 0.205-0.767 &  0.132 \\
\hline 

\end{tabular}
\end{center}
\end{table}

\subsubsection{Quasi-Static Feedforward Control}
For a desired tip reference $p_{ref} \in \mathbb{R}^{2\times1}$, the feedforward input $u_{FF}$ is obtained by solving the static inverse problem $p_{model}(u) = p_{ref}$. This is formulated as a nonlinear least-squares problem minimizing the task-space residual, $r(u)=p_{model}(u)-p_{ref}$, where $p_{model} \in \mathbb{R}^{2\times1}$ is the model tip position. A Gauss–Newton iteration is used, based on the local input–output sensitivity, $S(\theta)$, given by:
\begin{align} \label{eq:S}
    S(\theta) = \frac{\partial p_{model}}{\partial u}
    = J(\theta)K_{\theta}^{-1}T_{\theta} \in \mathbb{R}^{2 \times 2},   
\end{align}
where $T_{\theta} \in R^{N\times 2}$ maps actuator inputs to generalized model torques ($T_{\theta}(i,j)=\pm1$ if input $j$ results in a counterclockwise/clockwise torque on rod $i$ and is $0$ otherwise). The reduced rotational stiffness matrix $K_{\theta}\in \mathbb{R}^{N\times N}$ is constructed from the corresponding rotational elements of the full stiffness matrix $K \in \mathbb{R}^{3N \times 3N}$ and includes elastic coupling between adjacent rods. Each column of the Jacobian $J(\theta)=\frac{\partial p_{model}}{\partial \theta} \in \mathbb{R}^{2\times N}$ is given by $J_{i}= 
\begin{bmatrix}
    -L_i \sin{\theta_i} \\
    L_i \cos{\theta_i}
\end{bmatrix}$. 
Equation (\ref{eq:S}) was derived as follows: we considered the minimal set of coordinates $\theta = [\theta_1,...,\theta_N]^T$ and noted that for small perturbations about equilibrium $\delta \theta= K_{\theta}^{-1} T_{\theta} \delta u$ and $\delta p_{model} = J_{\theta} \delta \theta$ which leads to $\delta p_{model} = S(\theta) \delta u$. 
The update term at iteration $k$ is:

\begin{equation}
    \Delta u_{FF,k} =
    \begin{cases}
    -S^{-1}r(u_k), & \text{if } S \text{ is} \\
    & \text{well-conditioned,} \\
    -(S^T S + \epsilon I)^{-1}S^Tr(u_k), & \text{otherwise,}
\end{cases}
\end{equation}
where we abbreviate $S=S(\theta_k)$ for compactness, $\epsilon$ is the regularization parameter (1$\mathrm{e}{-8}$), and S($\theta_k$) is considered ill-conditioned if its reciprocal condition number is $<$1$\mathrm{e}{-10}$. The regularized form maintains robustness near kinematic or actuation singularities. The input is updated as:
\begin{equation}
    u_{FF,k+1} = u_{FF,k} + \alpha_u\Delta u_{FF,k},
\end{equation}
where the step size $\alpha_u \in (0,1]$ is selected via backtracking to ensure monotonic reduction of $||r(u_k)||_2^2$. This procedure is applied pointwise along the reference trajectory $p_{ref}(t)$ to generate the feedforward input sequence $u_{FF}(t)$.

Because this inverse problem is solved pointwise using the static equilibrium map, $u_{FF}$ should be interpreted as a quasi-static feedforward term rather than an inverse-dynamics input and does not explicitly compensate inertial terms such as $M\ddot q$. Dynamic tracking errors are addressed by the observer and task-space PI feedback described below.

\subsubsection{Dynamic observer}
Tip and base positions are measured via optical tracking and transformed into the task-space frame. These measurements are incorporated into a dynamic observer that integrates Eq. (\ref{eq:fullsystem}) with initial conditions set to the unactuated equilibrium configuration and parameters from Table \ref{tab:parameters}. We use a discretization of $N=10$ rods, which was found to provide a practical balance between modeling fidelity and real-time computational efficiency. Coarser discretizations reduce computation time but underrepresent distributed curvature variation, particularly for P2. We therefore selected a resolution higher than typical 3--5 segment PCC dynamic implementations, while still allowing the $N=10$ observer/controller loop to run in real time at 1 kHz.

State correction is achieved by injecting measurement residuals as generalized forces through the vector $Q$. Specifically, this virtual tip force is applied at the distal rod to steer the states of the observer toward the configuration of the physical actuator. This force, $Q_{obs} \in \mathbb{R}^{2 \times 1}$ enters as an external input to the $\Ddot{x}_N,\Ddot{y}_N$ dynamics (available to us from the non-minimal coordinate formulation) and is given by:
\begin{gather}
    Q_{obs} = K_{o}(p_{data} - p_{model}) + B_{o}(\dot{p}_{data} - \dot{p}_{model}).
\end{gather}
Here $p_{data}$ and $p_{model}$ are the measured and model tip positions and $K_{o},B_{o} \in \mathbb{R}^{2\times2}$ are the observer gain matrices. In all experiments, we used $K_o=\mathrm{diag}([100,\ 100])$ and $B_o=0$. The gain $K_o$ was tuned empirically by increasing it until the observer tip converged rapidly to the measured tip without excessive oscillation or noise amplification; this proportional correction was sufficient to obtain low observer error, so derivative correction was not used.

Pneumatic actuation dynamics were modeled as a first-order response (50 ms time constant) with approximately 50 ms pure delay and incorporated into the observer forward dynamics. Pressure characterization supports this approximation (Supplementary Section S1, Fig. S1). While this model does not fully capture the asymmetry between chamber pressurization and exhaust, the mismatch is treated as model error corrected by task-space feedback.

\subsubsection{Task-space PI control}
Feedback is implemented in task space by mapping a desired task-space effort ($f_{des}$) to actuator inputs via the sensitivity matrix $S(\theta)$, which is computed online from the dynamic observer states. The desired effort combines proportional–integral feedback with a velocity feedforward term:
\begin{align} \label{eq:fdes}
    f_{des} &= K_Pe + K_I \int e dt + K_{FF}\dot{p}_{ref}, \\
    e &= p_{ref} - p_{data},
\end{align}
where $K_P,K_I,K_{FF} \in \mathbb{R}^{2 \times 2}$ and $e \in \mathbb{R}^{2 \times 1}$. The velocity feedforward term was empirically found to be useful for dynamic trajectory following tasks (Task 2). The corresponding feedback correction is obtained through a regularized least-squares solution via $S(\theta)$:
\begin{gather}
    u_{FB} = (S(\theta)^TS(\theta) + \epsilon I)^{-1} S(\theta)^T f_{des}, \label{eq:vff}
\end{gather}
which ensures numerical robustness near singular configurations. The total control input is:
\begin{equation}
    u = u_{FF} + u_{FB}.
\end{equation}
This input is saturated and linearly mapped to pressure setpoints, which are regulated by low-level PI loops at each proportional spool valve. Note that only the feedback component is computed from the dynamic observer; the feedforward component remains a pointwise quasi-static inverse and does not perform inverse-dynamics compensation.

\begin{figure}
  \centering
  \includegraphics[width=\linewidth]{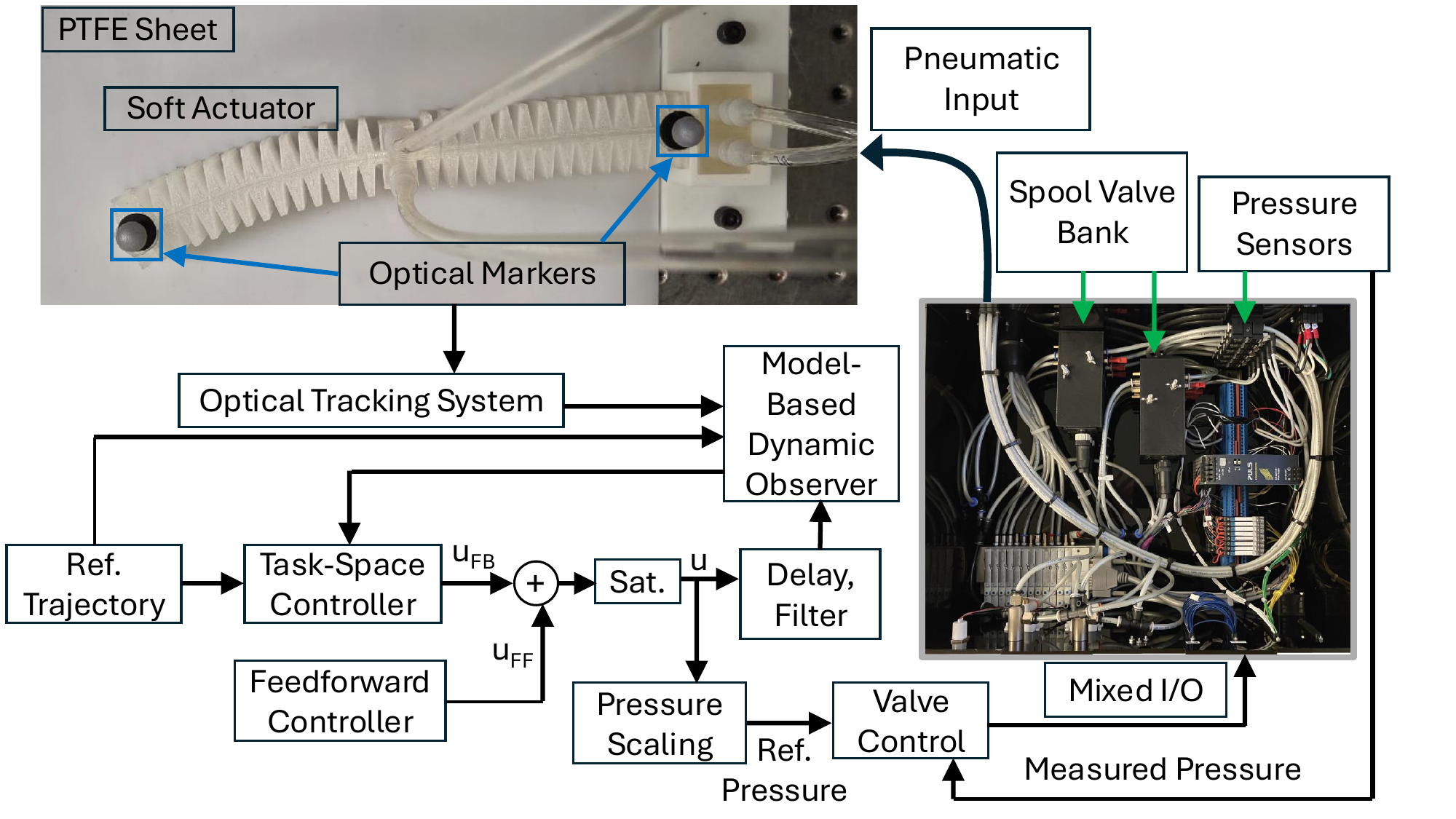}
  \caption{The system block diagram is shown here. The reference trajectory is processed by a quasi-static inverse (feedforward) module to generate $u_{FF}$. A dynamic observer integrates the constrained rod model and incorporates measurement residuals as virtual generalized forces to estimate the system state. The estimated state is used within the task-space PI control block to compute a feedback correction $u_{FB}$ via the local sensitivity mapping. The total input $u = u_{FF} + u_{FB}$ is applied through proportional valves with low-level pressure regulation.}  \label{fig:control}
\end{figure}

\subsection{Stability and Passivity Analysis}

The constrained rod model has a natural passive mechanical structure. For the nominal system, omitting Baumgarte stabilization terms, Eq. ({\ref{eq:fullsystem}}) may be written as:
\begin{equation} \label{eq:oneline}
    M\Ddot{q}+B\dot{q}+Kq = Tu + S_N^TQ_{obs}+A(q)^T\lambda_H,
\end{equation}
where $T \in \mathbb{R}^{3N\times2}$ is the zero-padded extension of $T_{\theta}$ and $S_N\in \mathbb{R}^{2 \times 3N}$ selects the translational coordinates of the distal rod such that $p_N = S_Nq$. Define the total mechanical energy:
\begin{equation}
    H(q,\dot{q}) = \frac{1}{2}\dot{q}^TM\dot{q} + \frac{1}{2}q^TKq.
\end{equation}
Taking the time derivative and substituting  Eq. ({\ref{eq:oneline}}):
\begin{equation}
    \dot{H} = -\dot{q}^TB\dot{q}+(T^T\dot{q})^Tu + \dot{p}_N^TQ_{obs} + \lambda_H^T A(q)\dot{q}.
\end{equation}
Since admissible velocities satisfy $A(q)\dot{q}=0$, holonomic constraint forces do no work. Therefore,
\begin{equation} \label{eq:Hdot}
    \dot{H} = -\dot{q}^TB\dot{q}+(T^T\dot{q})^Tu + \dot{p}_N^TQ_{obs}.
\end{equation}

Since $B\succeq0$, the discrete rod model is passive from the input-output pairs $u \mapsto T^T\dot{q}$ and $Q_{obs} \mapsto \dot{p}_N$. In the unforced case, $u=0$ and $Q_{obs}=0$, the energy derivative reduces to $\dot{H} = -\dot{q}^TB\dot{q} \leq 0$, so the nominal model is Lyapunov stable and dissipative. More generally, integrating Eq. ({\ref{eq:Hdot}}) gives:

\begin{equation}
    H(t) \leq H(0) + \int_0^t (T^T\dot{q})^T u d\tau + \int_0^t \dot{p}_N^T Q_{obs} d\tau.
\end{equation}

Thus, bounded supplied mechanical work from pneumatic actuation and observer force injection yields bounded stored mechanical energy. This analysis establishes passivity and dissipativity of the nominal constrained mechanical model used by the observer and controller, demonstrating that the underlying mechanical dynamics are not energy-generating. It does not constitute a stability proof of the complete implemented closed-loop system, which includes the observer, PI controller, pneumatic delays, pressure saturation, measurement noise, and numerical implementation.

\subsection{Experimental Platform}

\subsubsection{Soft Actuator}

The experimental platform consists of three planar pneumatic soft actuators (P1–P3) shown in Fig.~\ref{fig:pneunet}. P1 is used for the majority of the experiments, while P2 and P3 are used to evaluate generalizability of our approach. Each actuator follows the Pneunet architecture \cite{ilievski2011soft}, in which pressurization induces asymmetric expansion in compliant regions, producing planar bending. The actuator body contains four independent internal chambers, allowing bidirectional curvature control through differential pressurization.
All soft actuators were fabricated using a Bambu Lab X1C 3D printer with TPU (Polymaker PolyFlex TPU 90A). Actuator P1 was designed with a uniform wall thickness of 1.5 mm and 14 triangular segments per chamber, resulting in a total length of 253 mm. P2 shares the same overall geometry as P1 but features a spatially varying wall thickness, decreasing from 3 mm at the base to 1.1 mm at the tip. 
P3 is identical to P1 except that it contains 10 triangular segments per chamber, resulting in a reduced total length of 193 mm. All three actuators have a height of 27.5 mm.

\begin{figure}
  \centering
  \includegraphics[width=\linewidth]{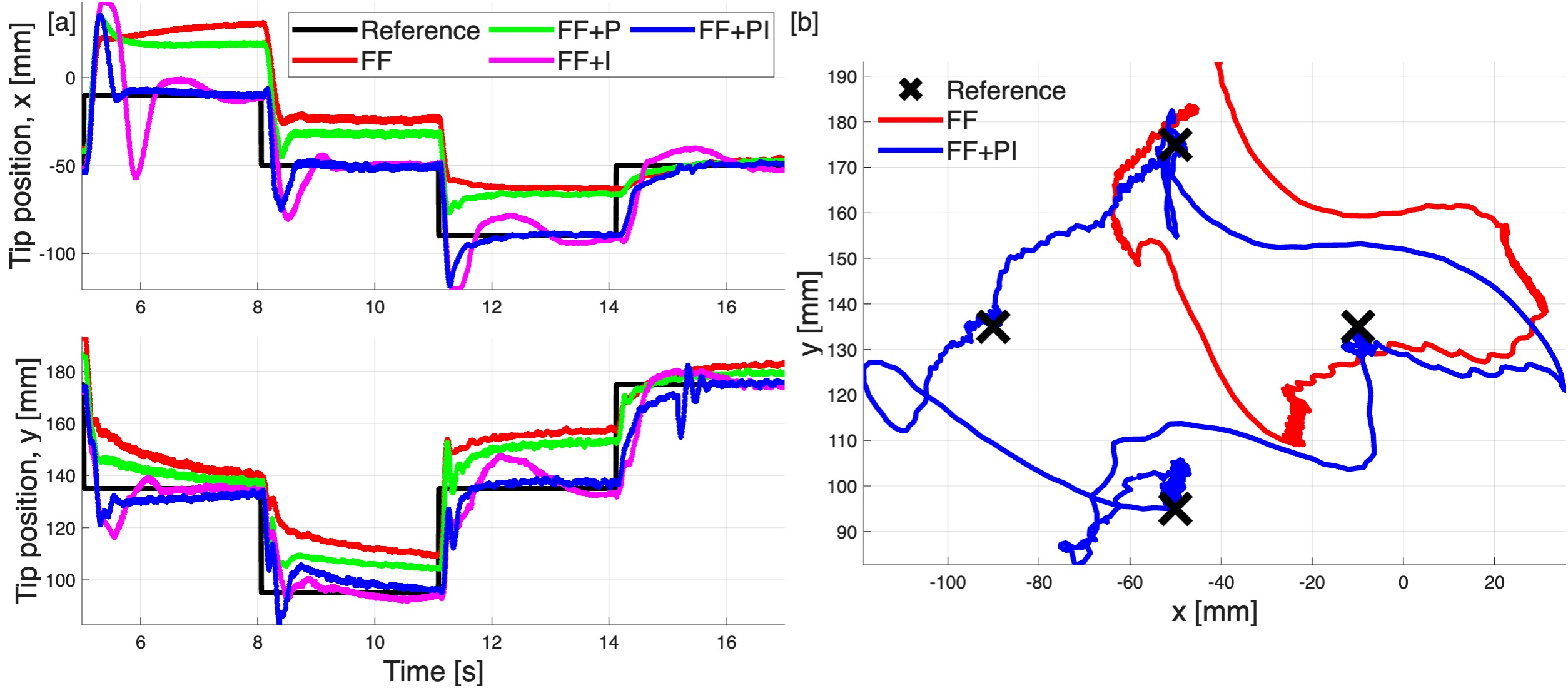}
  \caption{The temporal (a) and task-space (b) response of the soft actuator to four step inputs that draw a diamond in task-space. In (a), four controllers are used. The response of the feedforward (FF) and FF+PI are then shown in task-space (b). Black Xs represent each commanded step starting at the top and moving clockwise.}
  \label{fig:step}
\end{figure}

\subsubsection{Hardware setup}

Each of the four hollow chambers in each actuator is independently actuated using a custom pneumatic control box developed in our research group (Fig.~\ref{fig:control}). The chambers are connected via 4 mm ID, 6 mm OD pneumatic tubing (McMaster-Carr 50315K69) using barbed fittings (McMaster-Carr 5463K628). The soft actuator operates over a PTFE sheet (McMaster-Carr 9266K21) to reduce friction during operation and is secured to a mechanical breadboard using a custom 3D-printed fixture.

\begin{table}
\begin{center}
\caption{Step response average error characterization}
\label{tab:steperr}
\begin{tabular}{| c | c | c | c |}
\hline
Controller & RMSE [mm] & SSE [mm] & ISE [mm$^2$s] \\
\hline
FF & 30.5 & 15.7 & 3309 \\ \hline
FF+P & 23.8 & 11.3 & 2007 \\ \hline
FF+I & 19.1 & 1.0 & 1131 \\ \hline
FF+PI & 14.4 & 1.0 & 625 \\
\hline 
\end{tabular}
\end{center}
\end{table}

The control box contains four 5-port/3-way proportional spool valves (Festo MPYE-5-M5-010-B), each equipped with an inline pressure sensor (Festo SDE5-D10-NF-T14-V-M8). The system is interfaced through a mixed I/O data acquisition card (Humusoft MF634) to a desktop control computer (AMD Ryzen 7 5700G 3.80 GHz) running Matlab 2024b Simulink Desktop Real-Time. This configuration enables independent closed-loop pressure control of each chamber. Each chamber pressure was regulated using a low-level PI loop that converted pressure error into the corresponding proportional valve command. Pressure saturation was applied to enforce the actuator operating range and prevent self-collision.

Actuator motion is measured using an OptiTrack Prime 13 optical tracking system, with Cartesian position data streamed at 100 Hz. The controller, the numerical integration of Eq. (\ref{eq:fullsystem}) using Matlab's ode4 fixed step solver for the dynamic observer, pressure sensing, and spool valve command generation, are executed at 1 kHz. The mean computation time per step (as calculated by Simulink Profiler) was found to be 0.97 ms. 

\begin{figure}
  \centering
  \includegraphics[width=\linewidth]{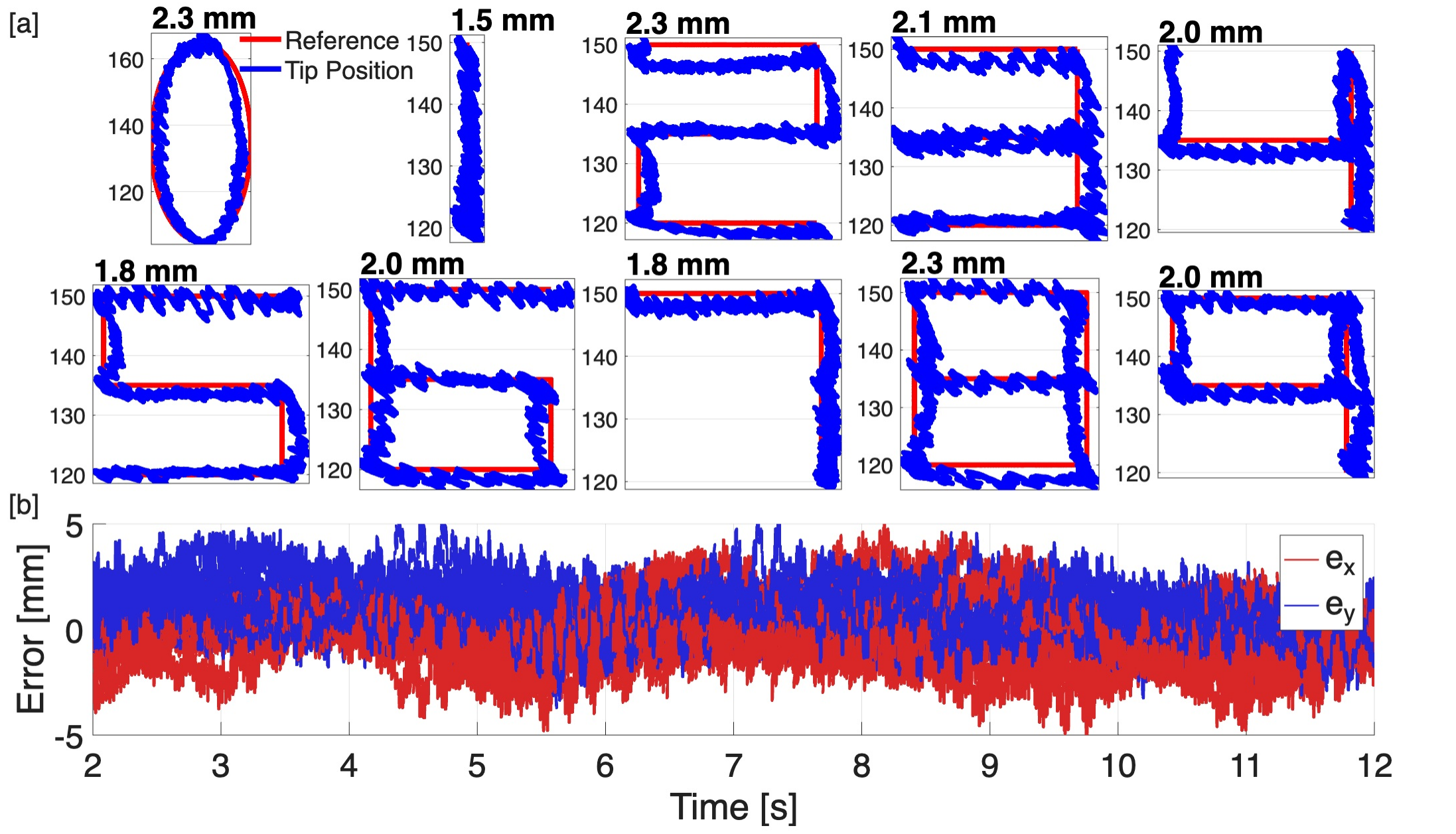}
  \caption{Task 1: The actuator tip is commanded to draw each of the 10 digits in 10 sec (a). The corresponding RMSE values are shown in the top left for each digit. The x and y errors ($e_x,e_y$) for all 10 plots are shown (b) and are always within 5 mm.}
  \label{fig:digits}
\end{figure}

\section{Experimental Results}

We begin by characterizing the step response of the soft actuator P1 under four control strategies: feedforward only (FF), FF with proportional (FF+P), FF with integral (FF+I), and FF+PI feedback. This was performed to demonstrate the relative importance of each component of the controller architecture. The actuator tip was commanded to a sequence of four task-space step inputs at 3 s intervals, forming a diamond-shaped trajectory (Fig.~\ref{fig:step}).

Since conventional step response metrics (e.g., rise time, overshoot) are not uniformly applicable across all responses, performance is evaluated using the integral squared error (ISE):
\begin{align}
    \text{ISE} (t)&= \int_{0}^{t} e(\tau)^2 d\tau,
\end{align}
which penalizes large transient deviations. We also report the average RMSE and steady-state error (SSE), calculated as the mean absolute error over the final 10$\%$ of each step interval. 

The results, summarized in Table~{\ref{tab:steperr}}, show that integral feedback (FF+I) was particularly important for reducing steady-state error, while the combination of proportional and integral feedback (FF+PI) provided the best overall transient and steady-state performance. FF+PI control was therefore used in all subsequent experiments, and its performance is evaluated across five representative tasks:

\begin{itemize}
\item \textbf{Task 1:} Precise tracking of slow trajectories (3–18 mm/s),
\item \textbf{Task 2:} High-speed and high-frequency tracking (19 cm/s, 1 Hz),
\item \textbf{Task 3:} Cross-platform generalization to actuators P2 and P3 (Fig.~\ref{fig:pneunet}a),
\item \textbf{Task 4:} Tip tracking using only intermediate backbone sensing, 
\item \textbf{Task 5:} Real-time user-defined trajectory tracking.
\end{itemize}

\subsection*{Task 1: Precise Tracking of Slow Trajectories}

Two experiments were conducted to evaluate precise low-speed tip tracking of P1. In the first experiment, the actuator tip was commanded to draw each of the ten digits over 10 s per digit. The reference trajectories consisted of a 60 mm × 30 mm ellipse for ``0'', a 30 mm vertical line for ``1'', and 30 mm × 30 mm square-bounded paths for ``2–9''. Due to variations in path length, reference speeds ranged from 3–18 mm/s.

The task-space references and tip trajectories are shown in Fig.~\ref{fig:digits}a, with the corresponding RMSE values indicated in each subplot. The reference trajectory starts from the unactuated equilibrium position and transitions to the beginning of each digit in 2 s. The error plots in Fig.~\ref{fig:digits}b show that the tracking error remains below 5 mm along each axis throughout the drawing task for all digits.

\begin{figure}
  \centering
  \includegraphics[width=\linewidth]{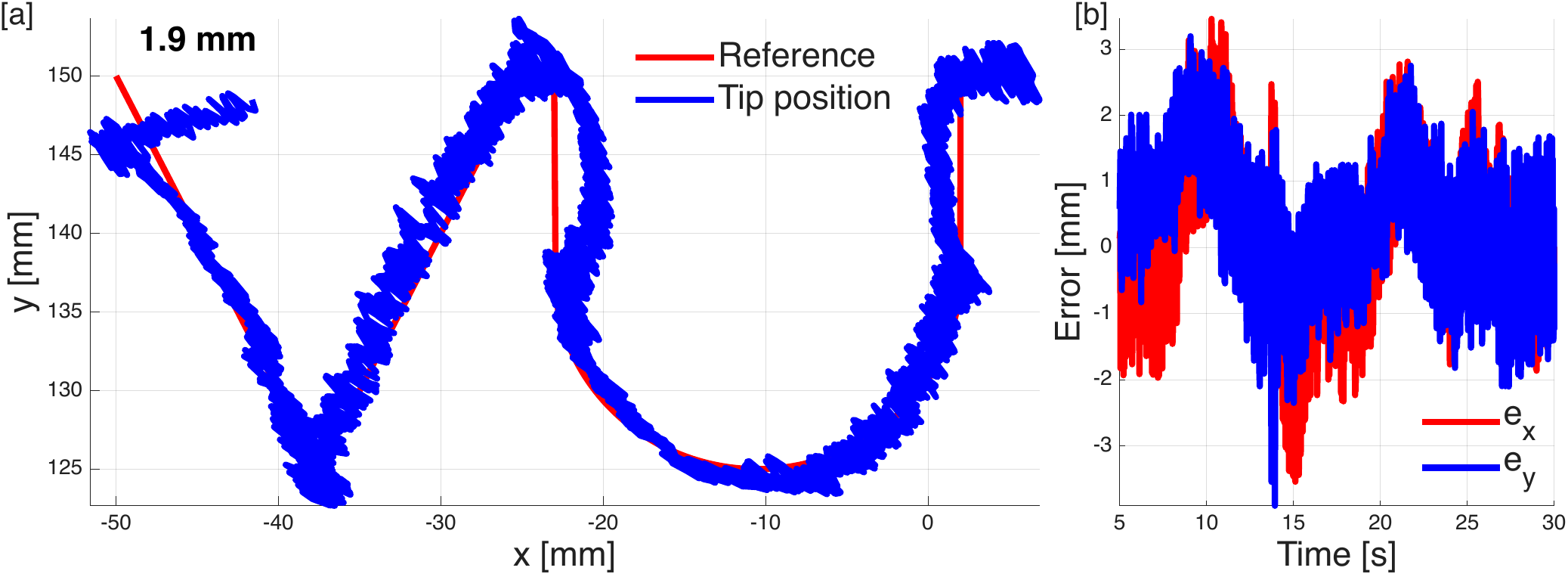}
  \caption{Task 1: The actuator tip draws the letters ``VU'' spanning 25 mm in height and 55 mm in width in 25 sec. This task was accomplished with errors less than 4 mm with an RMSE of 1.9 mm (b).}
  \label{fig:VU}
\end{figure}

The second experiment involves writing the letter ``VU'' over 25 s (Fig.~\ref{fig:VU}). The reference trajectory for this task ranges from x=-50 to 5 mm and from y=125 to 150 mm, spanning a 125 mm trajectory. The reference speed was held constant at 5 mm/s. The tracking error remained below 4 mm throughout the motion, with an RMSE of 1.9 mm. This task required substantial curvature variation, including motion toward the unactuated equilibrium configuration (y=0), further illustrating precision throughout task-space.

\subsection*{Task 2: High-speed tracking}

We next evaluate tracking performance under higher-speed and higher-frequency trajectories. Two reference motions were considered: a 60 mm diameter circular trajectory and a 60 mm vertical line trajectory at 1 Hz. In both cases, the actuator transitions from the unactuated configuration to the oscillatory reference by $t = 2$ s, followed by 7 s of periodic tracking.

\begin{figure}
  \centering
  \includegraphics[width=\linewidth]{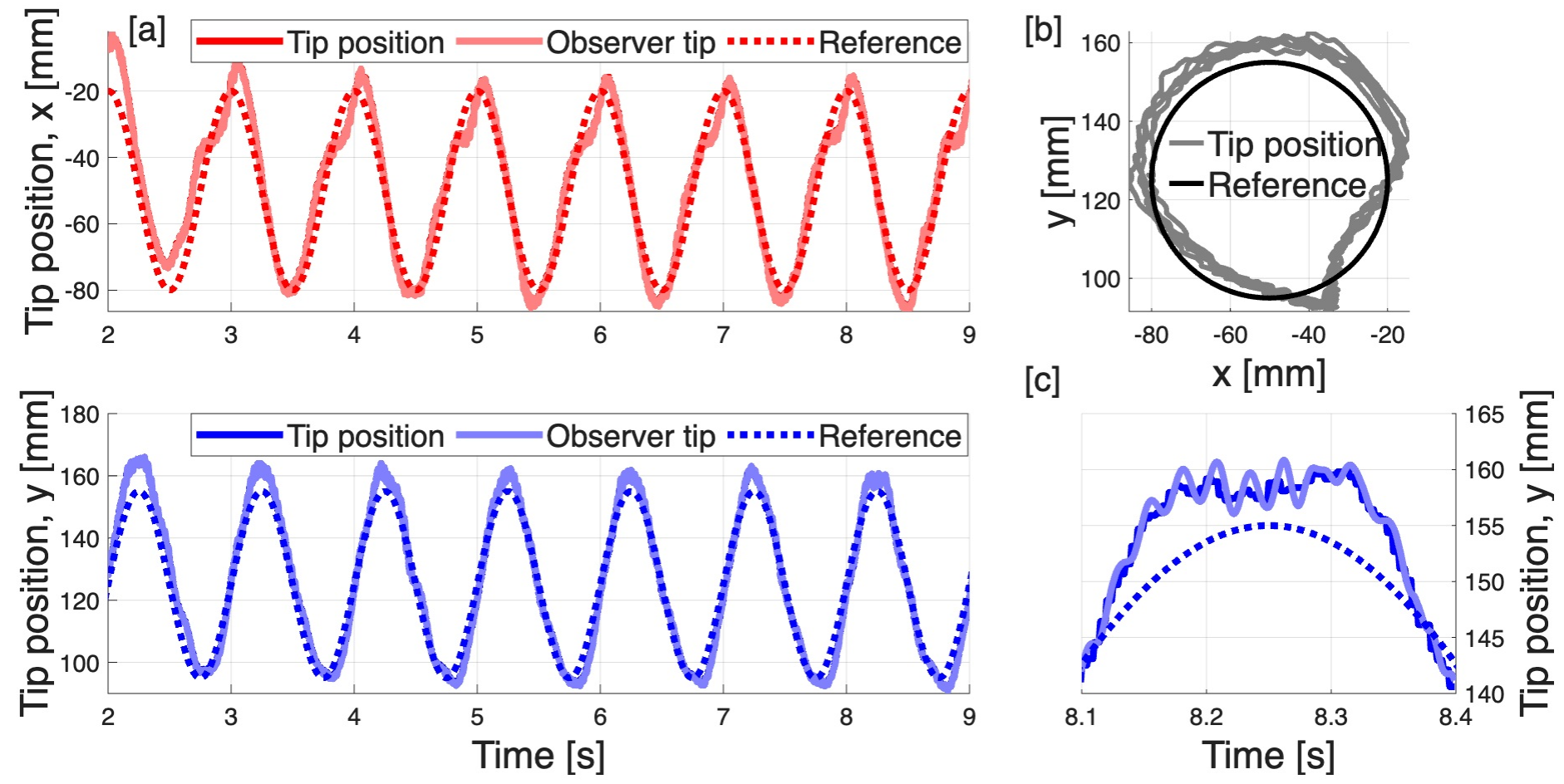}
  \caption{Task 2: The actuator is commanded to follow a 60 mm 1 Hz circle. The actual tip position, observer tip position, and reference for x and y are shown (a). The task-space trajectory for the last 4 cycles is shown (b). The observer tip closely tracks the actual tip position - a zoomed view of the y-position from t = 8.1-8.4 s is shown in (c).}
  \label{fig:1hzcircle}
\end{figure}

\begin{figure}
  \centering
  \includegraphics[width=\linewidth]{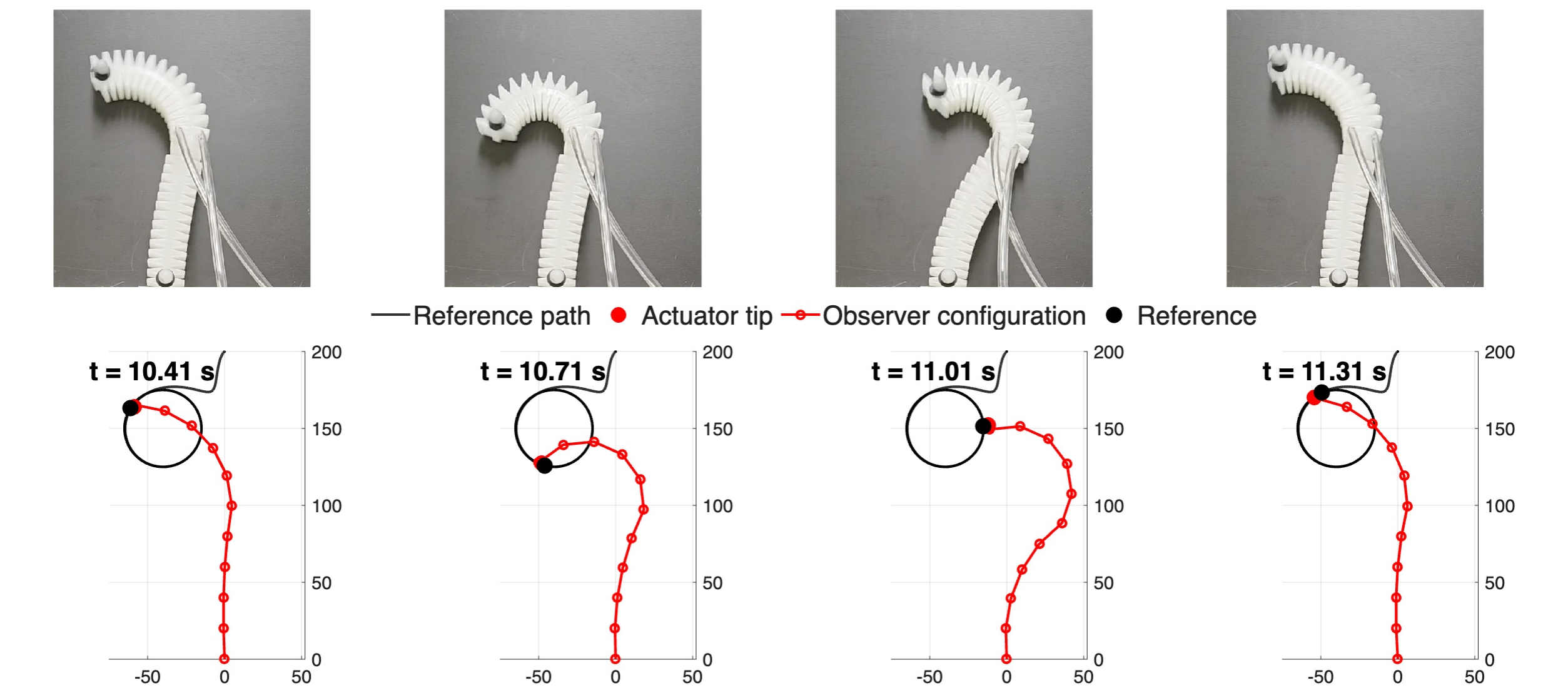}
  \caption{Task 2: Four snapshots of the actuator during one cycle of the 1 Hz circle-following task are shown (spaced 0.3 sec apart). Top row images are extracted from a phone camera and bottom row images depict the configuration of the observer during each instance.}
  \label{fig:1hzcirclesnap}
\end{figure}

Figure~\ref{fig:1hzcircle} presents results for the 60 mm, 1 Hz circular trajectory, corresponding to a peak tip speed of 18.8 cm/s and peak acceleration of 1.18 m/s$^2$. In Fig.~\ref{fig:1hzcircle}(a), the top and bottom plots show the x and y positions of the actuator tip position (data), observer, and reference. The task-space trajectory in Fig.~\ref{fig:1hzcircle}(b) demonstrates close agreement between the reference and measured motion over the final cycles. The zoomed view in Fig.~\ref{fig:1hzcircle}(c) highlights the close alignment between the observer and measured tip positions. Figure~\ref{fig:1hzcirclesnap} further compares the physical actuator and dynamic observer configurations at four time points within a cycle, illustrating consistency between the 10-link observer backbone and the actuator shape.

\begin{figure}
  \centering
  \includegraphics[width=\linewidth]{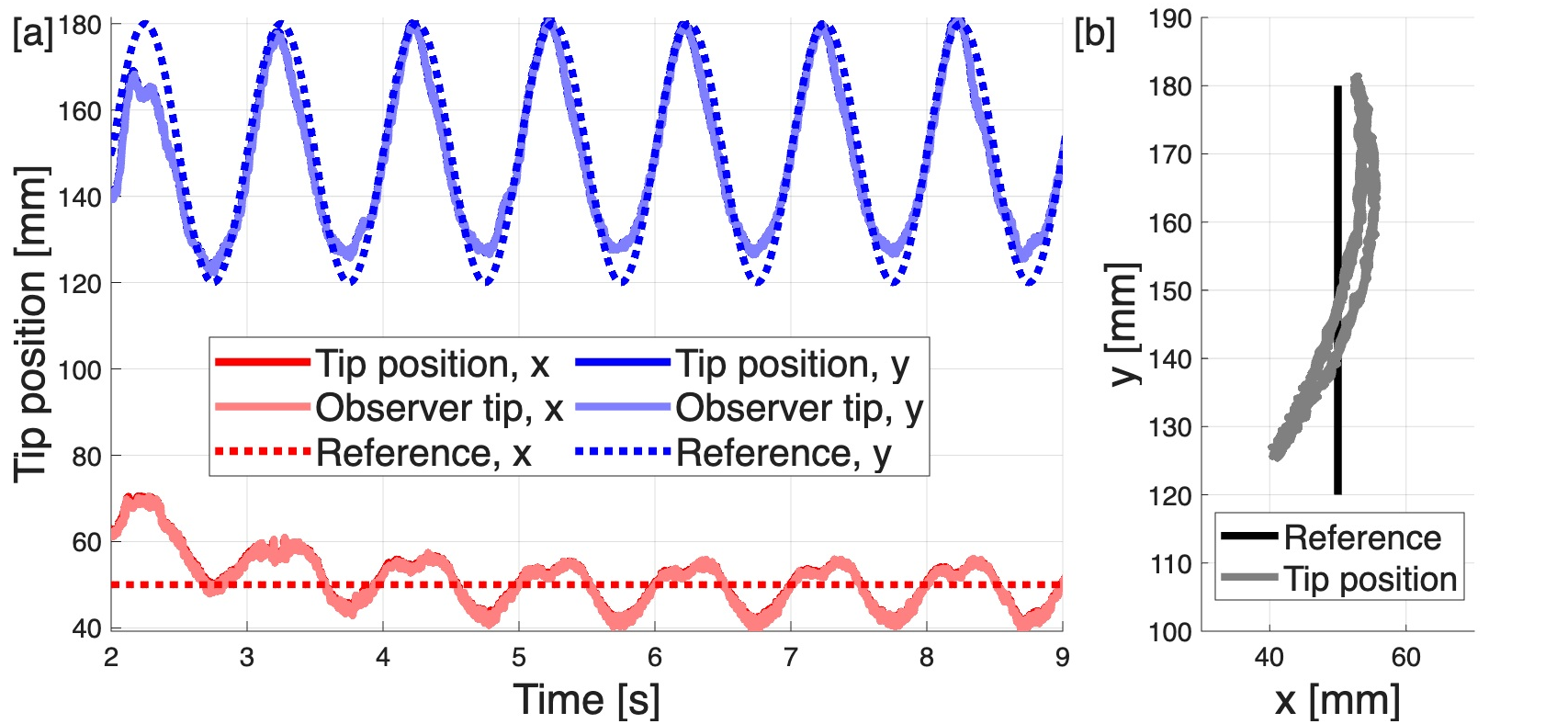}
  \caption{Task 2: The actuator is commanded to follow a 60 mm 1 Hz vertical line. The actual tip position, observer tip position, and reference for x and y are shown (a). The task-space trajectory for the last 4 cycles is shown (b).}
  \label{fig:1hzVL}
\end{figure}

Figure~\ref{fig:1hzVL} presents the 60 mm, 1 Hz vertical line trajectory. While tracking remains stable, larger deviations are observed compared to circular motion, particularly in the transverse direction. Enforcing linear task-space motion requires coordinated curvature variation that is more challenging for the bending-dominated actuator.

A summary of performance across Tasks 1 and 2 is provided in Table~\ref{tab:t12}. Here, the characteristic length refers to the nominal size of the reference trajectory (e.g., length, diameter). We also include a horizontal line following task at different frequencies. For slow trajectories (Task 1), millimeter-scale accuracy is achieved at tip speeds around 10 mm/s. For dynamic trajectories (Task 2), stable tracking is maintained up to approximately 19 cm/s with RMSE on the order of 7–12 mm, demonstrating moderate-bandwidth task-space control without instability or excessive phase lag.

\begin{table}
\begin{center}
\caption{Summary of Tasks 1 and 2.}
\label{tab:t12}
\begin{tabular}{| c | c | c | c | c |}
\hline
Task & Task & Characteristic & Characteristic & RMSE [mm]\\
No. &  & Length [mm] & Time &  \\
\hline
1 & Digits & 30 & 10 s & 1.5-2.3 \\
\hline
1 & VU & 50 & 25 s & 1.9 \\ 
\hline
2 & Vertical line & 60 & 0.25 Hz & 7.4 \\
& & & 1 Hz & 8.8 \\
\hline
2 & Horizontal line & 60 & 0.25 Hz & 9.2 \\
 &  & & 1 Hz & 12.4 \\
\hline 
2 & Circle & 60 & 1 Hz & 7.4 \\
\hline 
\end{tabular}
\end{center}
\end{table}

\subsection*{Task 3: Generalization to Varying Actuator Geometries}

\begin{figure}
  \centering
  \includegraphics[width=\linewidth]{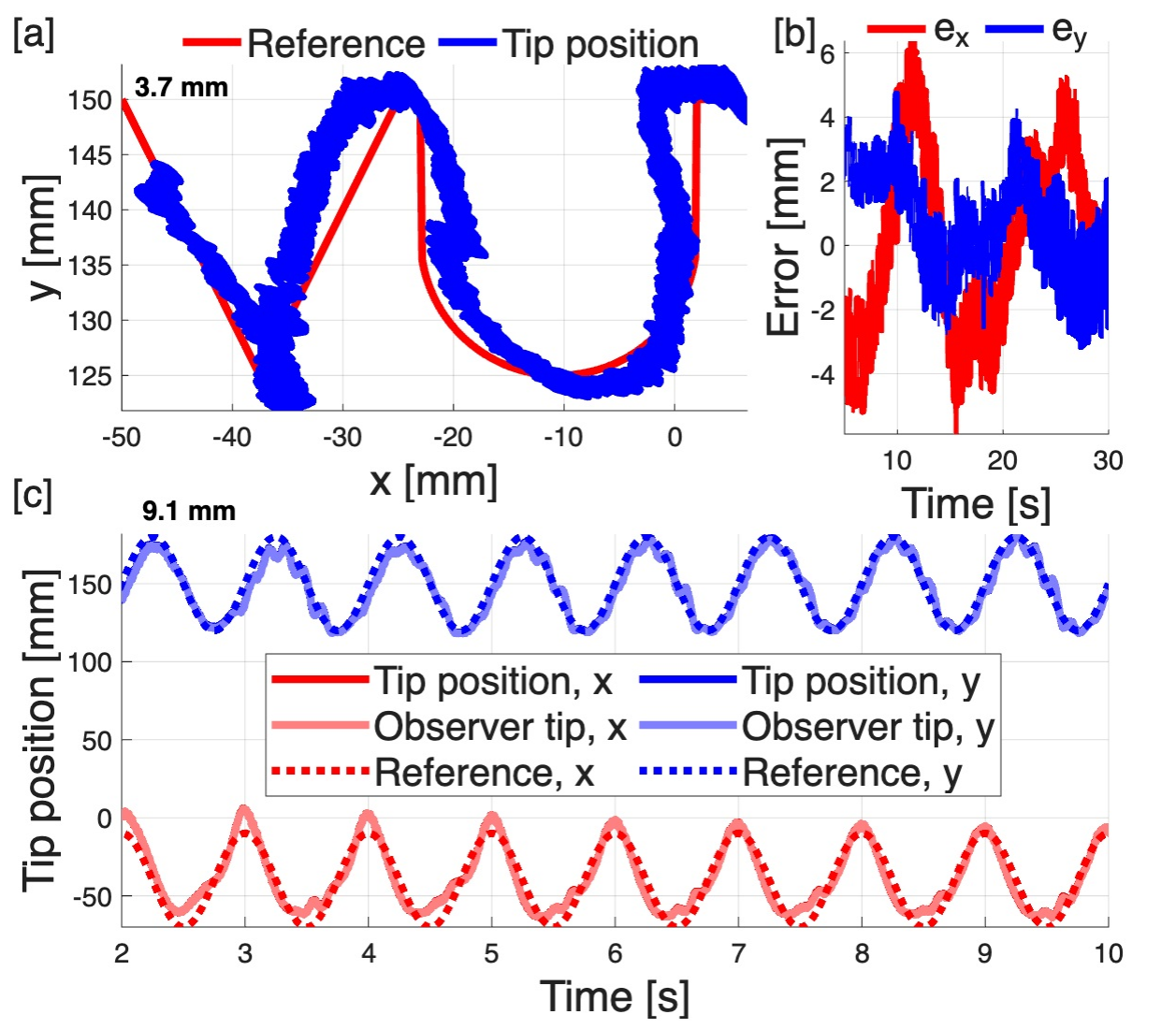}
  \caption{Task 3: The variable-curvature Pneunet (P2) is commanded to repeat Tasks 1 and 2. In (a), ``VU" is drawn in 25 s with an RMSE of 3.7 mm. The corresponding error plot is shown in (b). In (c), a 60 mm, 1 Hz circle is tracked with an RMSE of 9.1 mm.}
  \label{fig:spirob}
\end{figure}

To evaluate generalizability, one test each from Tasks 1 and 2 was repeated using two additional actuators, P2 and P3 (Fig.~\ref{fig:pneunet}(a)). Only model parameters were updated for each actuator (Table~\ref{tab:parameters}) with no structural modifications to the observer or controller. 

Figure~\ref{fig:spirob} summarizes the results for P2. Despite the spatially varying wall thickness and resulting nonuniform curvature distribution, the controller achieves an RMSE of 3.7 mm for the ``VU'' task and 9.1 mm for the 60 mm, 1 Hz circular trajectory, with no appreciable phase lag. 

\begin{figure}
  \centering
  \includegraphics[width=\linewidth]{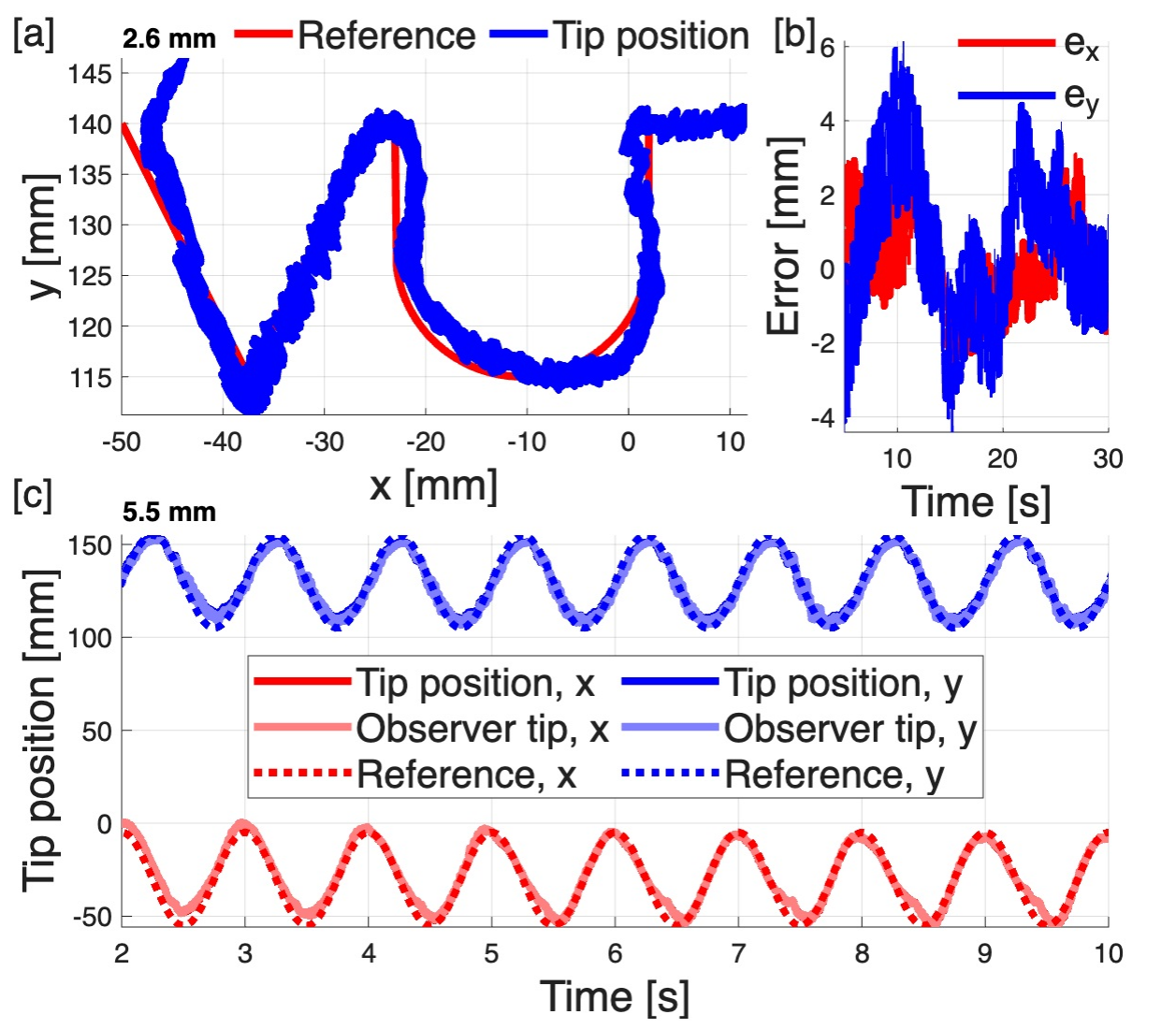}
  \caption{Task 3: The shorter Pneunet (P3) repeats Tasks 1 and 2. In (a), ``VU" is drawn in 25 s with an RMSE of 2.6 mm. The corresponding error plot is shown in (b). In (c), a 60 mm, 1 Hz circle is tracked with an RMSE of 5.5 mm. Due to the reduced workspace, trajectories are vertically shifted.}
  \label{fig:short}
\end{figure}

\begin{figure}
  \centering
  \includegraphics[width=\linewidth]{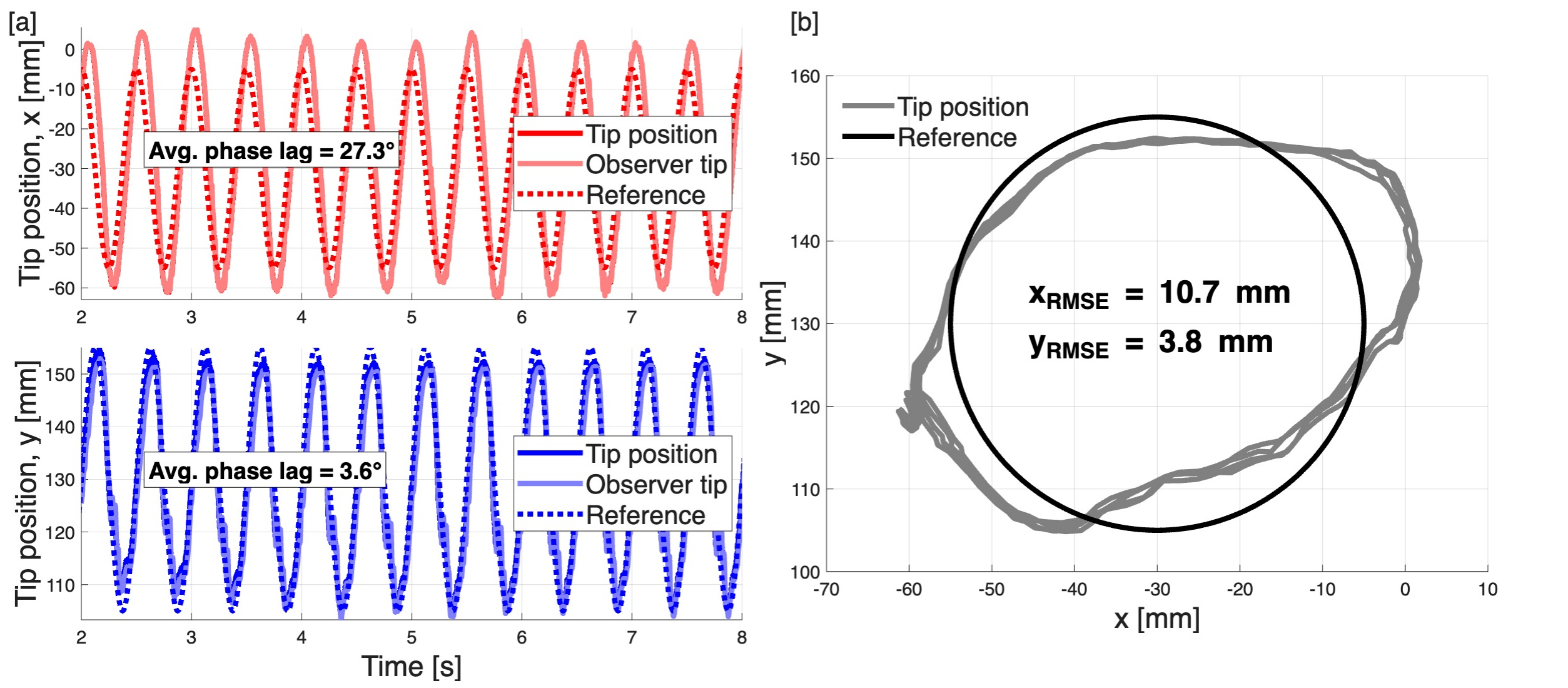}
  \caption{Task 3: The short Pneunet (P3) is able to track a 60 mm 2 Hz circle. The x and y tip positions along with the respective phase lags are shown (a). The task-space trajectory for the last 4 cycles and RMSE for the entire 6 sec run are shown (b).}
  \label{fig:short2hz}
\end{figure}

Figure~\ref{fig:short} presents results for P3, which differs from P1 primarily in overall length. To accommodate its reduced workspace, reference trajectories were vertically shifted without changing their size. The controller achieves RMSE values of 2.6 mm for the ``VU'' task and 5.5 mm for the 1 Hz circular trajectory. Encouraged by this performance, we further evaluated bandwidth limits using P3. A 60 mm, 2 Hz circular trajectory (37 cm/s peak speed, 4.7 m/s$^2$ peak acceleration) was successfully tracked (Fig.~\ref{fig:short2hz}). While tracking remained stable, increased phase lag ($27^\circ$ in the x-direction) and larger error (11.4 mm RMSE) were observed, reflecting the bandwidth limitations of the closed-loop architecture at higher frequency. These limitations arise from the combined effects of pneumatic delay, actuator compliance, finite feedback bandwidth, observer/model mismatch, and the quasi-static feedforward term. A supplementary frequency sweep (Supplementary Section S2) quantifies this trend: the mean corrective feedback contribution increased from 12$\%$ of the total commanded input at 0.1 Hz to 32$\%$ at 2 Hz.

\begin{figure}
  \centering
  \includegraphics[width=\linewidth]{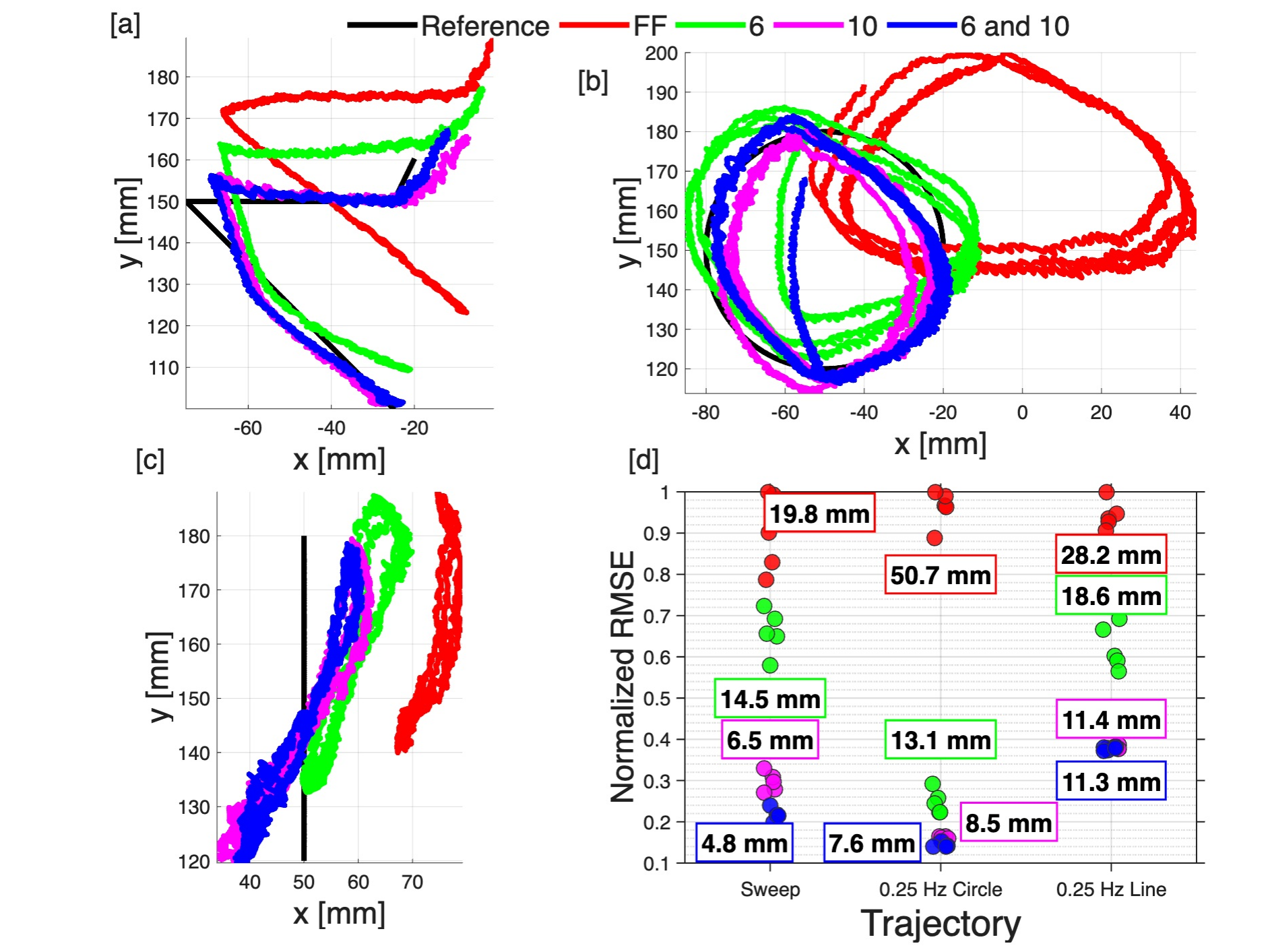}
  \caption{Task 4: Performance under four sensing conditions — open-loop (FF), rod 6 only, rod 10 only (tip sensing), and rods 6 and 10 combined. The three reference trajectories are (a) sweep, (b) 60 mm, 0.25 Hz circle, and (c) 60 mm, 0.25 Hz vertical line. Each condition is repeated five times, and the distribution of normalized tip RMSE is shown in (d). Average absolute RMSE values for each trajectory and sensing configuration are also reported.}
  \label{fig:T4}
\end{figure}

 \begin{figure}
  \centering
  \includegraphics[width=\linewidth]{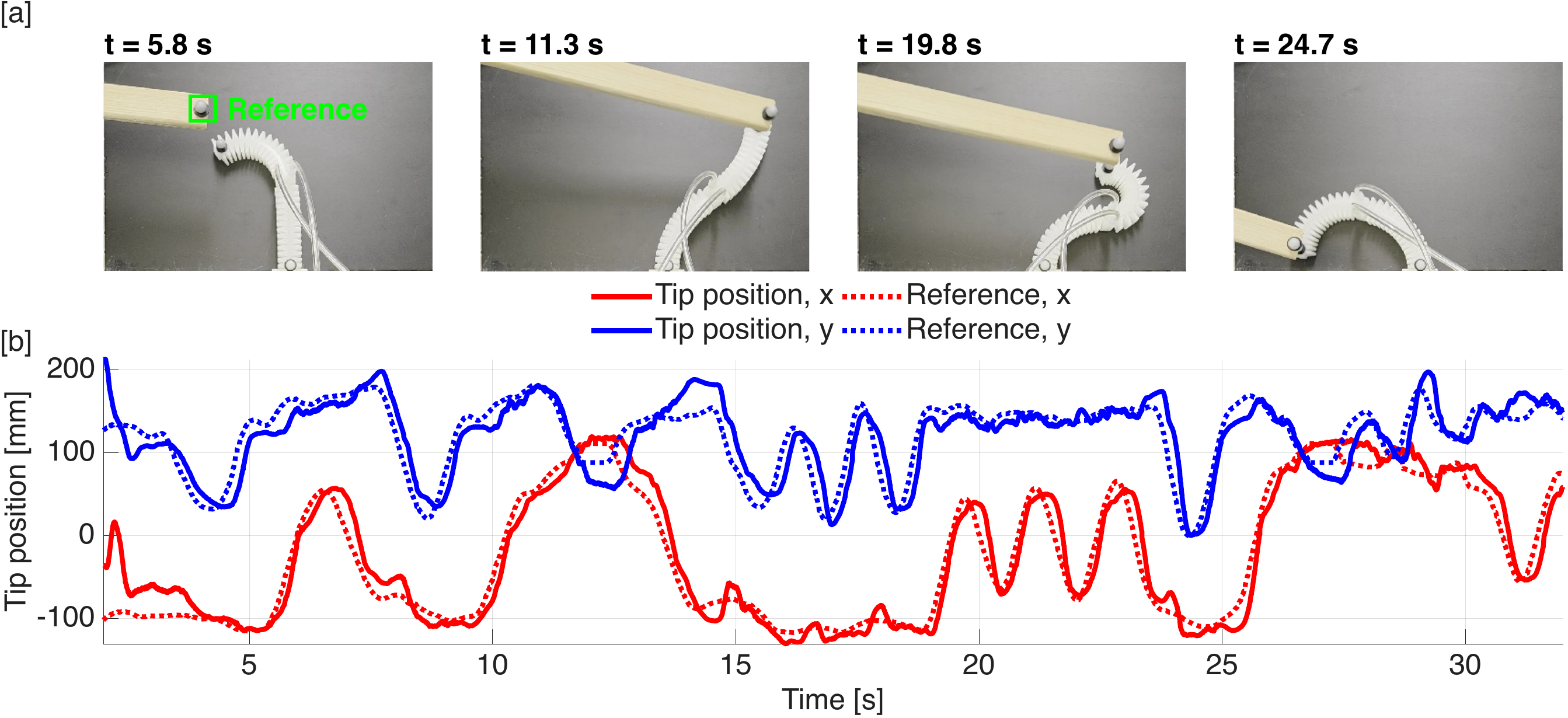}
  \caption{Task 5: Real-time tracking of an arbitrary user-defined reference generated by an optical marker attached to a handheld rod is demonstrated. Representative snapshots during the 30 s trial are shown in (a). The corresponding projected reference and actuator tip trajectories in x and y are shown in (b).}
  \label{fig:T5}
\end{figure}
  
\subsection*{Task 4: Reduced Sensing}

We next consider the case where direct tip measurements are unavailable and only an intermediate backbone position is accessible. To remain consistent with the observer discretization ($N=10$), we assume optical tracking is available only near rod 6 (with rod 1 at the base and rod 10 at the tip). Four sensing configurations are evaluated:
\begin{enumerate}
\item Open-loop (FF only),
\item Rod 6 only,
\item Rod 10 only (nominal sensing),
\item Rods 6 and 10 combined.
\end{enumerate}
The open-loop condition corresponds to feedforward control without feedback. The observer framework allows real-time fusion of position (and velocity) data from any selected rods; in the final configuration, positional data from both rods 6 and 10 are incorporated simultaneously.
Each sensing condition was tested on three reference trajectories:
\begin{enumerate}
\item Sweep,
\item 60 mm, 0.25 Hz circle,
\item 60 mm, 0.25 Hz vertical line.
\end{enumerate}
Each trajectory was repeated five times per sensing configuration. Figures~\ref{fig:T4}(a–c) show representative task-space trajectories (reference and all four sensing configurations): (a) sweep, (b) circle, and (c) vertical line. Figure~\ref{fig:T4}(d) summarizes the distribution of tip RMSE values. For each trajectory (x-axis), the distribution of tip position RMSE is shown for five runs per sensing condition. RMSE values are normalized by the maximum RMSE across all 20 trials per trajectory (four sensing conditions $\times$ five runs). The average absolute RMSE per trajectory is also shown. 

Incorporating intermediate backbone sensing significantly improves performance relative to open-loop control. Compared to FF, using the position data of rod 6 reduces RMSE by 27$\%$, 74$\%$, and 34$\%$ for the sweep, circle, and line trajectories, respectively. Combining rods 6 and 10 yields a slight but not statistically significant improvement over tip-only sensing.

\subsection*{Task 5: Real-Time User-Defined Tracking}

In this final experiment, we evaluate real-time tracking performance without a precomputed reference trajectory or feedforward input. The actuator tip is commanded to follow an optical marker affixed to the end of a handheld rod. Unlike previous tasks, the reference is generated online and varies continuously and arbitrarily over time. Three modifications to the control strategy are introduced:
\begin{enumerate}
\item The measured marker position is projected onto the actuator’s planar workspace and used as the task-space reference.
\item The offline quasi-static feedforward component is replaced with an online inverse update computed at a lower rate (100 Hz), while the observer and pressure control loops remain at 1 kHz.
\item The velocity feedforward term in Eq. (\ref{eq:fdes}) is disabled.
\end{enumerate}
Figure~\ref{fig:T5}(a) provides representative snapshots during a 30 s trial in which the reference marker is moved dynamically throughout the workspace. The projected reference and measured tip trajectories are shown in Fig.~\ref{fig:T5}(b), demonstrating stable real-time tracking under non-periodic and user-defined motion.

\section{Discussion}

This work positions non-minimal coordinate discrete elastic rod models as a practical alternative to both reduced-order PCC formulations and learning-dominant control approaches for soft pneumatic actuators. The results suggest a shift in the prevailing modeling–control tradeoff: real-time task-space control with dynamic model-based state estimation does not require either minimal-coordinate reduction or heavy data-driven augmentation when the mechanics are structured appropriately.

\subsection{Comparison with Prior Soft-Actuator Tracking Results}

Table~\ref{tab:controlpapers} summarizes representative results from prior dynamic control studies of soft continuum manipulators. Much of the literature has focused on curvature tracking, quasi-static task-space control, or dynamic tracking at relatively low bandwidths. PCC-based implementations commonly operate below 0.5 Hz with cm-level Cartesian accuracy, while learning-based approaches often report 10–30 mm errors over multi-second dynamic tasks. Recent Koopman-based and residual modeling strategies achieve 5–70 mm errors at frequencies up to approximately 0.5–1.1 Hz, typically on longer or higher-DoF platforms.

In contrast, this work demonstrates mm-scale Cartesian accuracy (1.5–2.3 mm RMSE) for slow trajectories and sub-cm accuracy (5.5–12.4 mm RMSE) for periodic tracking at 1–2 Hz. To our knowledge, this represents one of the highest reported closed-loop task-space tracking bandwidths for pneumatically actuated soft continuum robots while maintaining mm-level precision. 

\begin{table*}
\caption{Recent literature summary of dynamic control of soft continuum manipulators. \\FBL=Feedback Linearization, PP=Pole Placement, NARX = Nonlinear Autoregressive with Exogenous Inputs, UKF=Unscented Kalman Filter, AP=Adaptive Passivity control, ANCF=Abs. Nodal Coordinate Formulation \label{tab:controlpapers}}
\centering
\begin{tabular}{| p{0.075\linewidth} | p{0.1\linewidth} | p{0.1\linewidth} | p{0.15\linewidth} | p{0.05\linewidth} | p{0.35\linewidth} |}
\hline
\textbf{Ref. (Year)} & \textbf{Model} & \textbf{Controller(s)} & \textbf{Platform} & \textbf{Size} & \textbf{Free-space tracking results} \\
\hline

\cite{Falkenhahn2017} (2017) & 
3-link PCC & FBL & 
Festo Bionic Handling Assistant & 87 cm & 
$<$3 mm static and $\approx$50 mm dynamic error (0.5 Hz)  \\ 
\hline

\cite{Gillespie2018} (2018)  & 
NN & MPC & 
1-DOF pneumatic joint & N/A & 
2$\degree$ error to 40$\degree$ step angle tracking with 1–2 s rise time \\
\hline

\cite{DellaSantina2018} (2018) &  
5-link PCC & Curvature PID & 
Planar soft pneumatic actuator & 32 cm & 
5$\degree$ $L^2$ norm error (0.33 Hz) \\ 
\hline

\cite{Thuruthel2019} (2019) & 
Learned NARX model & Model-based RL policy & 
2-segment pneumatic soft actuator & 30 cm & 
26 mm Cartesian error (dynamic reaching) \\ 
\hline

\cite{Xavier2022} (2022)  &  
Lumped 2nd-order & FBL, PP/LQR, UKF & 
1-DOF Pneunet & 10 cm & 
6–10$\degree$ avg. error to angle staircase input (0.25-0.5 Hz) \\ 
\hline

\cite{Haggerty2023} (2023) & 
Koopman & LQR & 
2 PMAs & 37 cm, 53 cm & 
5–10 mm error (from reported plots) (1.1 Hz, 1.52 m/s) \\ 
\hline

\cite{Bruder2025} (2025)  & 
Koopman & MPC & 
3-segment PMA & 72 cm & 
67 mm RMSE (0.1 Hz circular tracking)  \\ 
\hline

\cite{Huang2025} (2025) &
ANCF + NN & FBL &
3-DOF soft parallel robot & 17.5 cm &
0.4–1.1 mm tracking error (5–80 mm/s) \\ 
\hline

\cite{Azizkhani2025} (2025)  & 
3-link PCC & AP & 
9-DOF pneumatic actuator & 45 cm & 
11–43 mm $L^2$ norm error (0.5 Hz, 0.42 m/s) \\ 
\hline

\textbf{This work} &
Discrete elastic rod & FF + PI &
2-DOF planar Pneunet & 25 cm &
1.5–2.3 mm RMSE (slow tracking); 5.5–12.4 mm RMSE (1–2 Hz, 0.37 m/s) \\
\hline

\end{tabular}
\end{table*}

\subsubsection{Comparison to PCC-Based Dynamic Models}

PCC formulations remain the dominant modeling framework for continuum robots due to their geometric interpretability and reduced state dimension \cite{Webster2010a,Godage2016,Katzschmann2019}. However, this reduction introduces two structural limitations: (i) dense inertia matrices with cubic scaling in segment count \cite{Till2019,Boyer2021}, and (ii) reliance on curvature shape assumptions that limit the ability to capture higher-order deformation without increasing segment resolution. Consequently, most dynamic PCC implementations operate with 3–5 segments \cite{DellaSantina2018,Kapadia2014,Falkenhahn2017}, trading fidelity for real-time tractability.

In contrast, the non-minimal formulation adopted here preserves absolute coordinates with holonomic constraints while maintaining block-sparse structure. This enables real-time integration with 10 segments, capturing distributed curvature variation without imposing predefined shape constraints. The 1–2 Hz tracking results illustrate that stable dynamic task-space control can be achieved without quasi-static curvature assumptions. Thus, here we demonstrate that non-minimal coordinates can invert the traditional tractability–fidelity tradeoff when sparsity is exploited correctly.

\subsubsection{Comparison to Learning-Based Control Approaches}

Learning-based strategies, ranging from neural inverse models to MPC with learned dynamics, have demonstrated strong empirical performance in capturing nonlinear behavior \cite{Thuruthel2019,Bruder2025,Haggerty2023}. However, these methods introduce structural tradeoffs: performance depends on training data coverage, extrapolation across actuator geometries or operating regimes may be limited, and stability guarantees are typically empirical.

In the present framework, generalization across P1, P2, and P3 required only parameter updates, with no retraining or structural controller modification. Physical constraints are enforced directly through the Lagrangian formulation rather than learned implicitly. For pneumatically actuated continuum systems of this class, structured model-based dynamics can achieve high-precision, moderate-bandwidth task-space control without learned inverse mappings.

\subsection{Structural Advantages and Real-Time Adaptability}

\subsubsection{Observer-Based State Estimation}

Soft robots often rely on dense sensing (e.g., embedded strain sensors or full-field vision) to compensate for modeling uncertainty \cite{Laschi2016,Katzschmann2019}. In contrast, this work demonstrates that a virtual force-injected dynamic observer can reconstruct full-state behavior from sparse Cartesian measurements alone.

The intermediate-point sensing experiment (Task 4) is particularly instructive. Even without direct tip sensing, meaningful trajectory tracking was achieved (average 45\% reduction in RMSE relative to FF only). This indicates that the dynamic structure itself contributes to observability. Such model-consistent observability is difficult to replicate in purely learning-driven approaches without dense sensing or recurrent architectures.

This property has practical implications for soft robotic deployment. In many real-world tasks such as grasping, insertion, environmental interaction, or operation in cluttered spaces, the actuator tip may be occluded or physically inaccessible to sensors. In such scenarios, reconstructing tip behavior from proximal or intermediate measurements is essential. The ability to maintain stable task-space tracking using only backbone sensing suggests that structured dynamic models can reduce sensing requirements while preserving control performance. This lowers hardware complexity and improves robustness in settings where visual or embedded tip sensing is unreliable or unavailable.

The observer residual should be interpreted as an aggregate correction term rather than a physically identified external wrench. In free-space tracking, this is useful because errors due to parameter mismatch, pneumatic delay, and unmodeled dissipation are converted into stabilizing virtual forces that keep the model state consistent with sparse measurements. However, in contact-rich tasks, the same residual may also contain contributions from environmental contact. With only position measurements, the observer cannot uniquely distinguish contact forces from internal model error or actuation mismatch. Therefore, the present implementation should not be interpreted as a contact-force observer. Decoupling these effects would require additional information, such as pressure/force sensing, known environmental/contact constraints, multiple distributed measurements, or an augmented disturbance/contact estimator. Contact-rich manipulation and force regulation are therefore outside the scope of this study and are left for future work.

\subsubsection{Real-Time Adaptability and Arbitrary Tracking}

Task 5 demonstrates real-time tracking of an online, user-defined reference without a precomputed feedforward trajectory. The actuator followed a handheld optical marker over a large portion of the workspace (approximately 250 mm in x and over 200 mm in y - the actuator itself is 253 mm in length). Tracking remained stable under aggressive, non-periodic excitation, although a measurable phase lag was observed due to pneumatic delay, the absence of the velocity feedforward term, and the reduced online inverse-update rate. This result suggests that the structured dynamic formulation can support responsive user-defined tracking without retraining or model restructuring.

To our knowledge, real-time tracking of arbitrary, user-generated trajectories using a purely mechanics-based dynamic model (without learning, residual correction, or policy adaptation) has rarely been demonstrated for pneumatically actuated continuum systems. Prior work using Koopman-based linear controllers \cite{Haggerty2023} reported similar dynamic excitation under learned models. Here, comparable behavior is achieved using only a structured dynamic formulation and sparse sensing.

This capability is significant for practical deployment. Many soft robotic applications such as human–robot interaction, assistive manipulation, teleoperation, and exploratory tasks require responsive, real-time adaptation to unpredictable references. Demonstrating stable arbitrary tracking without retraining or model restructuring suggests that structured non-minimal dynamics can serve as a reliable backbone for such applications.

\subsection{Limitations}

First, the present study is restricted to planar modeling and control. Although recent work has extended the underlying formulation to spatial 3D rods using a symmetric constrained-Lagrangian formulation with quaternions and Baumgarte stabilization \cite{Gaston2026}, the experimental validation here does not include torsional effects or out-of-plane dynamics. In that 3D formulation, a six-link 3D dynamic observer using backward Euler integration (0.025 s step) reduced mean position error from 7.6\% to 1.9\% of robot length, demonstrating real-time feasibility for moderate discretizations. Extending the present controller to full spatial Cosserat rod models is therefore computationally plausible, but achieving the 1 kHz update rate used in this planar implementation will require additional optimization and experimental validation.

Because the actuators are 3D printed from TPU, material hysteresis, fabrication variability, and fatigue can alter the effective stiffness and damping over time. The present implementation uses offline-identified parameters and does not explicitly adapt $k_i$ and $b_i$ online. Moderate parameter mismatch is partially corrected by task-space feedback and observer residual injection, but prolonged systematic drift may reduce observer accuracy, particularly under reduced-sensing conditions. Online parameter adaptation, periodic recalibration, or actuator replacement may mitigate this limitation.

Finally, the experiments presented here validate planar free-space task-space tracking. Although the underlying constrained-Lagrangian framework accommodates both holonomic and non-holonomic constraints \cite{Kumar2026}, the present evaluation does not include contact interaction, force regulation, or manipulation. Likewise, full 3D motion, torsional dynamics, and long-term adaptation to material degradation remain to be experimentally validated.

\section{Conclusion}

This work demonstrates that non-minimal coordinate discrete elastic rod models can be used for real-time planar free-space task-space control of soft pneumatic continuum actuators. Across five experimental tasks—including precision tracking, high-speed motion, actuator generalization, reduced sensing, and real-time user-driven references—the proposed controller architecture achieved stable and accurate performance without relying on reduced-order curvature assumptions or learning-based inverse mappings. More broadly, the results show that when sparsity is exploited, constrained non-minimal formulations can retain distributed mechanical fidelity while remaining computationally viable for planar free-space control. Future work will extend this framework to spatial rod models and evaluate contact-rich manipulation, further establishing physics-grounded structured dynamics as a scalable foundation for high-performance soft robotic control.

\section*{Funding Data}

This work was funded by a Vanderbilt University Seeding Success grant.

\section*{Conflicts of Interest}

The authors declare no conflicts of interest.

\bibliographystyle{ieeetr}
\bibliography{ref_trim}

@ARTICLE{Burgner-Kahrs2015,
  author={Burgner-Kahrs, Jessica and Rucker, D. Caleb and Choset, Howie},
  journal={IEEE Transactions on Robotics}, 
  title={Continuum Robots for Medical Applications: A Survey}, 
  year={2015},
  volume={31},
  number={6},
  pages={1261-1280},
  doi={10.1109/TRO.2015.2489500}}

@ARTICLE{DupontSurvey2022,
  author={Dupont, Pierre E. and Simaan, Nabil and Choset, Howie and Rucker, D. Caleb},
  journal={Proceedings of the IEEE}, 
  title={Continuum Robots for Medical Interventions}, 
  year={2022},
  volume={110},
  number={7},
  pages={847-870}}

@article{Laschi2016,
author = {Cecilia Laschi  and Barbara Mazzolai  and Matteo Cianchetti },
title = {Soft robotics: Technologies and systems pushing the boundaries of robot abilities},
journal = {Science Robotics},
volume = {1},
number = {1},
pages = {eaah3690},
year = {2016},
doi = {10.1126/scirobotics.aah3690},
URL = {https://www.science.org/doi/abs/10.1126/scirobotics.aah3690},
eprint = {https://www.science.org/doi/pdf/10.1126/scirobotics.aah3690}}

@article{ilievski2011soft,
author = {Ilievski, Filip and Mazzeo, Aaron D. and Shepherd, Robert F. and Chen, Xin and Whitesides, George M.},
title = {Soft Robotics for Chemists},
journal = {Angewandte Chemie International Edition},
volume = {50},
number = {8},
pages = {1890-1895},
doi = {https://doi.org/10.1002/anie.201006464},
url = {https://onlinelibrary.wiley.com/doi/abs/10.1002/anie.201006464},
eprint = {https://onlinelibrary.wiley.com/doi/pdf/10.1002/anie.201006464},
year = {2011}
}

@article{El-Atab2020,
author = {El-Atab, Nazek and Mishra, Rishabh B. and Al-Modaf, Fhad and Joharji, Lana and Alsharif, Aljohara A. and Alamoudi, Haneen and Diaz, Marlon and Qaiser, Nadeem and Hussain, Muhammad Mustafa},
title = {Soft Actuators for Soft Robotic Applications: A Review},
journal = {Advanced Intelligent Systems},
volume = {2},
number = {10},
pages = {2000128},
doi = {https://doi.org/10.1002/aisy.202000128},
url = {https://advanced.onlinelibrary.wiley.com/doi/abs/10.1002/aisy.202000128},
eprint = {https://advanced.onlinelibrary.wiley.com/doi/pdf/10.1002/aisy.202000128},
year = {2020}
}

@Article{Su2022,
AUTHOR = {Su, Hang and Hou, Xu and Zhang, Xin and Qi, Wen and Cai, Shuting and Xiong, Xiaoming and Guo, Jing},
TITLE = {Pneumatic Soft Robots: Challenges and Benefits},
JOURNAL = {Actuators},
VOLUME = {11},
YEAR = {2022},
NUMBER = {3},
ARTICLE-NUMBER = {92},
URL = {https://www.mdpi.com/2076-0825/11/3/92},
ISSN = {2076-0825},
DOI = {10.3390/act11030092}
}

@ARTICLE{Xavier2022b,
  author={Xavier, Matheus S. and Tawk, Charbel D. and Zolfagharian, Ali and Pinskier, Joshua and Howard, David and Young, Taylor and Lai, Jiewen and Harrison, Simon M. and Yong, Yuen K. and Bodaghi, Mahdi and Fleming, Andrew J.},
  journal={IEEE Access}, 
  title={Soft Pneumatic Actuators: A Review of Design, Fabrication, Modeling, Sensing, Control and Applications}, 
  year={2022},
  volume={10},
  number={},
  pages={59442-59485},
  doi={10.1109/ACCESS.2022.3179589}}

@article{Webster2010a,
author = {Robert J. Webster and Bryan A. Jones},
title ={Design and Kinematic Modeling of Constant Curvature Continuum Robots: A Review},
journal = {The International Journal of Robotics Research},
volume = {29},
number = {13},
pages = {1661-1683},
year = {2010},
doi = {10.1177/0278364910368147},

URL = { https://doi.org/10.1177/0278364910368147},
eprint = {https://doi.org/10.1177/0278364910368147}
}

@INPROCEEDINGS{Bailly2005,
  author={Bailly, Y. and Amirat, Y.},
  booktitle={Proceedings of the 2005 IEEE International Conference on Robotics and Automation}, 
  title={Modeling and Control of a Hybrid Continuum Active Catheter for Aortic Aneurysm Treatment}, 
  year={2005},
  volume={},
  number={},
  pages={924-929},
  doi={10.1109/ROBOT.2005.1570235}}

@InProceedings{Camarillo2009P,
author="Camarillo, David B.
and Carlson, Christopher R.
and Salisbury, J. Kenneth",
editor="Khatib, Oussama
and Kumar, Vijay
and Pappas, George J.",
title="Task-Space Control of Continuum Manipulators with Coupled Tendon Drive",
booktitle="Experimental Robotics",
year="2009",
publisher="Springer Berlin Heidelberg",
address="Berlin, Heidelberg",
pages="271--280",
isbn="978-3-642-00196-3"
}

@ARTICLE{Mahvash2011,
  author={Mahvash, Mohsen and Dupont, Pierre E.},
  journal={IEEE Transactions on Robotics}, 
  title={Stiffness Control of Surgical Continuum Manipulators}, 
  year={2011},
  volume={27},
  number={2},
  pages={334-345},
  doi={10.1109/TRO.2011.2105410}}

@article{DupontDesignTRO10,
author = {Dupont, Pierre E and Lock, Jesse and Itkowitz, Brandon and Butler, Evan},
journal = {IEEE Transactions on Robotics},
pages = {209--225},
title = {{Design and Control of Concentric-Tube Robots}},
volume = {26},
year = {2010}
}

@INPROCEEDINGS{DellaSantina2018,
  author={Della Santina, Cosimo and Katzschmann, Robert K. and Bicchi, Antonio and Rus, Daniela},
  booktitle={2018 IEEE International Conference on Soft Robotics (RoboSoft)}, 
  title={Dynamic control of soft robots interacting with the environment}, 
  year={2018},
  volume={},
  number={},
  pages={46-53},
  doi={10.1109/ROBOSOFT.2018.8404895}}

@ARTICLE{Xavier2022,
  author={Xavier, Matheus S. and Fleming, Andrew J. and Yong, Yuen Kuan},
  journal={IEEE/ASME Transactions on Mechatronics}, 
  title={Nonlinear Estimation and Control of Bending Soft Pneumatic Actuators Using Feedback Linearization and UKF}, 
  year={2022},
  volume={27},
  number={4},
  pages={1919-1927},
  doi={10.1109/TMECH.2022.3155790}}

@ARTICLE{Azizkhani2025,
  author={Azizkhani, Milad and Kousik, Shreyas and Chen, Yue},
  journal={IEEE Robotics and Automation Letters}, 
  title={Dynamic Task Space Control of Redundant Pneumatically Actuated Soft Robot}, 
  year={2025},
  volume={10},
  number={6},
  pages={6408-6415},
  doi={10.1109/LRA.2025.3568316}}

@ARTICLE{Braganza2007,
  author={Braganza, David and Dawson, Darren M. and Walker, Ian D. and Nath, Nitendra},
  journal={IEEE Transactions on Robotics}, 
  title={A Neural Network Controller for Continuum Robots}, 
  year={2007},
  volume={23},
  number={6},
  pages={1270-1277},
  doi={10.1109/TRO.2007.906248}}

@INPROCEEDINGS{Gillespie2018,
  author={Gillespie, Morgan T. and Best, Charles M. and Townsend, Eric C. and Wingate, David and Killpack, Marc D.},
  booktitle={2018 IEEE International Conference on Soft Robotics (RoboSoft)}, 
  title={Learning nonlinear dynamic models of soft robots for model predictive control with neural networks}, 
  year={2018},
  volume={},
  number={},
  pages={39-45},
  doi={10.1109/ROBOSOFT.2018.8404894}}

@article{Haggerty2023,
author = {David A. Haggerty  and Michael J. Banks  and Ervin Kamenar  and Alan B. Cao  and Patrick C. Curtis  and Igor Mezic and Elliot W. Hawkes },
title = {Control of soft robots with inertial dynamics},
journal = {Science Robotics},
volume = {8},
number = {81},
pages = {eadd6864},
year = {2023},
doi = {10.1126/scirobotics.add6864},
URL = {https://www.science.org/doi/abs/10.1126/scirobotics.add6864},
eprint = {https://www.science.org/doi/pdf/10.1126/scirobotics.add6864}}

@article{Godage2016,
  title={Dynamics for variable length multisection continuum arms},
  author={Godage, Isuru S and Medrano-Cerda, Gustavo A and Branson, David T and Guglielmino, Emanuele and Caldwell, Darwin G},
  journal={The International Journal of Robotics Research},
  volume={35},
  number={6},
  pages={695--722},
  year={2016},
  publisher={SAGE Publications Sage UK: London, England}
}

@INPROCEEDINGS{Katzschmann2019,
  author={Katzschmann, Robert K. and Santina, Cosimo Della and Toshimitsu, Yasunori and Bicchi, Antonio and Rus, Daniela},
  booktitle={2019 2nd IEEE International Conference on Soft Robotics (RoboSoft)}, 
  title={Dynamic Motion Control of Multi-Segment Soft Robots Using Piecewise Constant Curvature Matched with an Augmented Rigid Body Model}, 
  year={2019},
  volume={},
  number={},
  pages={454-461}}

@article{Till2019,
author = {John Till and Vincent Aloi and D. Caleb Rucker},
title ={Real-time dynamics of soft and continuum robots based on Cosserat rod models},
journal = {The International Journal of Robotics Research},
volume = {38},
number = {6},
pages = {723-746},
year = {2019}}

@ARTICLE{Boyer2021,
  author={Boyer, Frederic and Lebastard, Vincent and Candelier, Fabien and Renda, Federico},
  journal={IEEE Transactions on Robotics}, 
  title={Dynamics of Continuum and Soft Robots: A Strain Parameterization Based Approach}, 
  year={2021},
  volume={37},
  number={3},
  pages={847-863}}

@INPROCEEDINGS{Kapadia2014,
  author={Kapadia, Apoorva D. and Fry, Katelyn E. and Walker, Ian D.},
  booktitle={2014 IEEE/RSJ International Conference on Intelligent Robots and Systems}, 
  title={Empirical investigation of closed-loop control of extensible continuum manipulators}, 
  year={2014},
  volume={},
  number={},
  pages={329-335},
  doi={10.1109/IROS.2014.6942580}}

@ARTICLE{Falkenhahn2017,
  author={Falkenhahn, Valentin and Hildebrandt, Alexander and Neumann, Rdiger and Sawodny, Oliver},
  journal={IEEE/ASME Transactions on Mechatronics}, 
  title={Dynamic Control of the Bionic Handling Assistant}, 
  year={2017},
  volume={22},
  number={1},
  pages={6-17},
  doi={10.1109/TMECH.2016.2605820}}

@ARTICLE{Thuruthel2019,
  author={Thuruthel, Thomas George and Falotico, Egidio and Renda, Federico and Laschi, Cecilia},
  journal={IEEE Transactions on Robotics}, 
  title={Model-Based Reinforcement Learning for Closed-Loop Dynamic Control of Soft Robotic Manipulators}, 
  year={2019},
  volume={35},
  number={1},
  pages={124-134},
  doi={10.1109/TRO.2018.2878318}}

@ARTICLE{Renda2020,
  author={Renda, Federico and Armanini, Costanza and Lebastard, Vincent and Candelier, Fabien and Boyer, Frederic},
  journal={IEEE Robotics and Automation Letters}, 
  title={A Geometric Variable-Strain Approach for Static Modeling of Soft Manipulators With Tendon and Fluidic Actuation}, 
  year={2020},
  volume={5},
  number={3},
  pages={4006-4013}}

@inproceedings{Rucker2022,
  title={Task-Space Control of Continuum Robots using Underactuated Discrete Rod Models},
  author={Rucker, Daniel Caleb and Barth, Eric J and Gaston, Joshua and Gallentine, James C},
  booktitle={2022 IEEE/RSJ International Conference on Intelligent Robots and Systems (IROS)},
  pages={10967--10974},
  year={2022},
  organization={IEEE}
}

@article{Kim2026,
    author = {Kim, Sung Y. and Kumar, Nithin S. and Barth, Eric J.},
    title = {Variable Stiffness and Impedance Control of a Soft Pneumatic Actuator With Tunable Collision Dynamics and Trajectory Following},
    journal = {Journal of Dynamic Systems, Measurement, and Control},
    volume = {148},
    number = {3},
    pages = {031003},
    year = {2025},
    month = {12},
    issn = {0022-0434},
    doi = {10.1115/1.4070266},
    url = {https://doi.org/10.1115/1.4070266},
    eprint = {https://asmedigitalcollection.asme.org/dynamicsystems/article-pdf/148/3/031003/7560221/ds-25-1141.pdf},
}

@article{Gaston2026,
    author = {Gaston, Joshua B. and Kumar, Nithin S. and Barth, Eric J. and Rucker, D. Caleb},
    title = {A 3D Discrete Elastic Rod Model and Observer for Continuum Robots},
    journal = {Journal of Mechanisms and Robotics},
    volume = {18},
    number = {3},
    pages = {031002},
    year = {2026},
    month = {01},
    issn = {1942-4302},
    doi = {10.1115/1.4070712},
    url = {https://doi.org/10.1115/1.4070712},
    eprint = {https://asmedigitalcollection.asme.org/mechanismsrobotics/article-pdf/18/3/031002/7579361/jmr-25-1182.pdf},
}

@article{Kumar2026,
    author = {Kumar, Nithin S. and Peters, John E. and Kim, Sung Y. and Gaston, Joshua and C. Rucker, D. and Barth, Eric J.},
    title = {Constrained Lagrangian-Based, Nonholonomic Modeling and Validation of a Pneumatically Actuated Wheeled Soft Robot},
    journal = {Journal of Mechanisms and Robotics},
    volume = {18},
    number = {4},
    pages = {041002},
    year = {2026},
    month = {03},
    issn = {1942-4302},
    doi = {10.1115/1.4071062},
    url = {https://doi.org/10.1115/1.4071062},
    eprint = {https://asmedigitalcollection.asme.org/mechanismsrobotics/article-pdf/18/4/041002/7590300/jmr-25-1183.pdf},
}

@article{baumgarte,
title = {Stabilization of Constraints and Integrals of Motion in Dynamical Systems},
journal = {Computer Methods in Applied Mechanics and Engineering},
volume = {1},
number = {1},
pages = {1-16},
year = {1972},
issn = {0045-7825},
doi = {https://doi.org/10.1016/0045-7825(72)90018-7},
url = {https://www.sciencedirect.com/science/article/pii/0045782572900187},
author = {J. Baumgarte}
}

@INPROCEEDINGS{Katzschmann2019v2,
  author={Katzschmann, Robert K. and Thieffry, Maxime and Goury, Olivier and Kruszewski, Alexandre and Guerra, Thierry-Marie and Duriez, Christian and Rus, Daniela},
  booktitle={2019 2nd IEEE International Conference on Soft Robotics (RoboSoft)}, 
  title={Dynamically Closed-Loop Controlled Soft Robotic Arm using a Reduced Order Finite Element Model with State Observer}, 
  year={2019},
  volume={},
  number={},
  pages={717-724}}

@article{Huang2020,
  title={Dynamic Simulation of Articulated Soft Robots},
  author={Huang, Weicheng and Huang, Xiaonan and Majidi, Carmel and Jawed, M. Khalid},
  journal={Nature Communications},
  volume={11},
  number={1},
  pages={2233},
  year={2020},
  publisher={Nature Publishing Group UK London}
}

@article{Bruder2025,
author = {Daniel Bruder and David Bombara and Robert J Wood},
title ={A Koopman-based residual modeling approach for the control of a soft robot arm},
journal = {The International Journal of Robotics Research},
volume = {44},
number = {3},
pages = {388-406},
year = {2025},
doi = {10.1177/02783649241272114},
URL = { https://doi.org/10.1177/02783649241272114},
eprint = {        https://doi.org/10.1177/02783649241272114
}
}

@ARTICLE{Huang2025,
  author={Huang, Xinjia and Rong, Yu and Gu, Guoying},
  journal={IEEE/ASME Transactions on Mechatronics}, 
  title={High-Precision Dynamic Control of Soft Robots With the Physics-Learning Hybrid Modeling Approach}, 
  year={2025},
  volume={30},
  number={3},
  pages={1658-1669},
  doi={10.1109/TMECH.2024.3403151}}

\section{Biography Section}

\begin{IEEEbiographynophoto}{Nithin S. Kumar} (Student Member, IEEE) received the B.S. degree (Vanderbilt University), the S.M. degree (MIT), and is currently a Ph.D. candidate in Mechanical Engineering at Vanderbilt University. His research interests include the design, modeling, and control of soft/continuum robots.  
\end{IEEEbiographynophoto}

\begin{IEEEbiographynophoto}{Joshua Gaston} (Student Member, IEEE) received the B.S. degree (Tennessee Tech University) and the M.S. degree (UTK), and is currently a Ph.D. candidate in Mechanical Engineering at UTK. His research interests include the design and control of soft/continuum robots.
\end{IEEEbiographynophoto}

\begin{IEEEbiographynophoto}{D. Caleb Rucker} (Senior Member, IEEE) is a B. Ray Thompson Professor in Mechanical Engineering at the University of Tennessee, Knoxville, where he directs the REACH lab. His research interests include design, modeling, and control of medical robots and soft/continuum robots. 
\end{IEEEbiographynophoto}

\begin{IEEEbiographynophoto}{Eric J. Barth} (ASME Fellow, IEEE Member) is a Professor in Mechanical Engineering at Vanderbilt University. His research interests include the design, modeling, and control of fluid power systems, soft and medical robotics.
\end{IEEEbiographynophoto}

\vfill

\end{document}